\documentclass[11pt,a4paper,twoside,openright]{report}

\usepackage{graphics}
\usepackage{color}
\usepackage{tabularx}
\usepackage{subfigure}
\usepackage{afterpage}
\usepackage{amsmath,amssymb}            
\usepackage{rotating}  
\usepackage{fancyhdr}  
\usepackage[scriptsize]{caption} 
\usepackage{url}
\usepackage{tabularx}
\usepackage{enumerate}
\usepackage{listings}
\usepackage{multirow}
\usepackage{float}
\newcommand\tabhead[1]{\small\textbf{#1}}
\newcommand{\degree}{\ensuremath{^\circ}}

\begin{document}
\setlength{\parindent}{0in}
\thispagestyle{empty}
{\large \textbf{POLITECNICO DI MILANO}}\\
\textbf{Scuola di Ingegneria dell'Informazione}\\
\begin{figure}[H]
	\includegraphics[width=3.5cm]{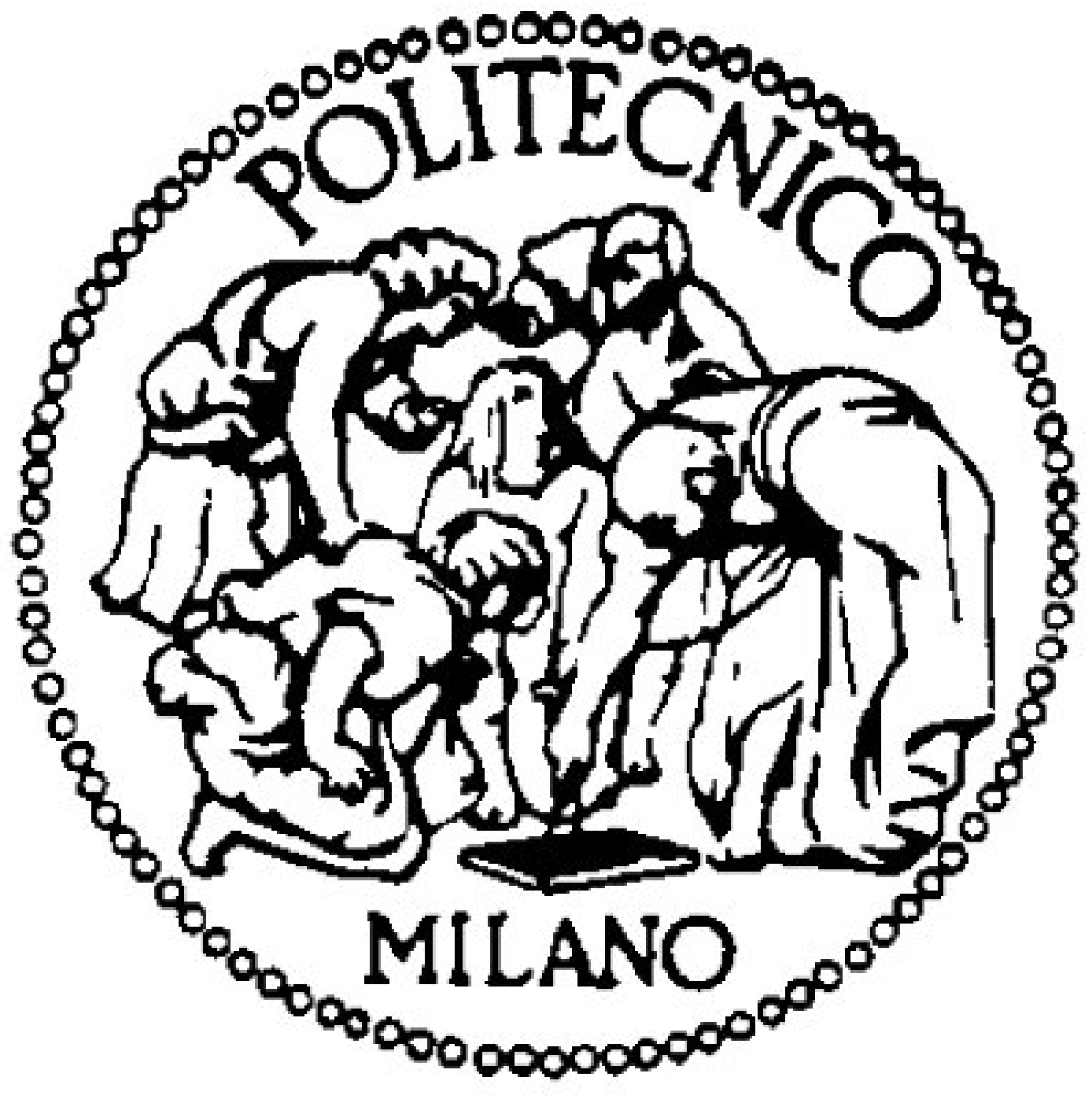}
\end{figure}
{\large \textbf{POLO TERRITORIALE DI COMO}}\\
{\small \textbf{Master of Science in Computer Engineering}}\\[1.5cm]
\linespread{1.6}
{\huge \textbf{Mountain Peak Detection\\ in Online Social Media\\ }}\\[1.0cm] 
\linespread{1}
\begin{flushleft}
{\bfseries Supervisor:} Prof. Marco Tagliasacchi\\
{\bfseries Assistant Supervisor:} Prof. Piero Fraternali\\[1.5cm]
\end{flushleft}
{\bfseries Master Graduation Thesis by:}\\
Roman Fedorov\\
id. 782110\\
\begin{flushright}
{\bfseries Academic Year} 2012/13
\end{flushright}
\thispagestyle{empty} \normalfont \cleardoublepage
\setlength{\parindent}{0in}
\thispagestyle{empty}
{\large \textbf{POLITECNICO DI MILANO}}\\
\textbf{Scuola di Ingegneria dell'Informazione}\\
\begin{figure}[H]
	\includegraphics[width=3.5cm]{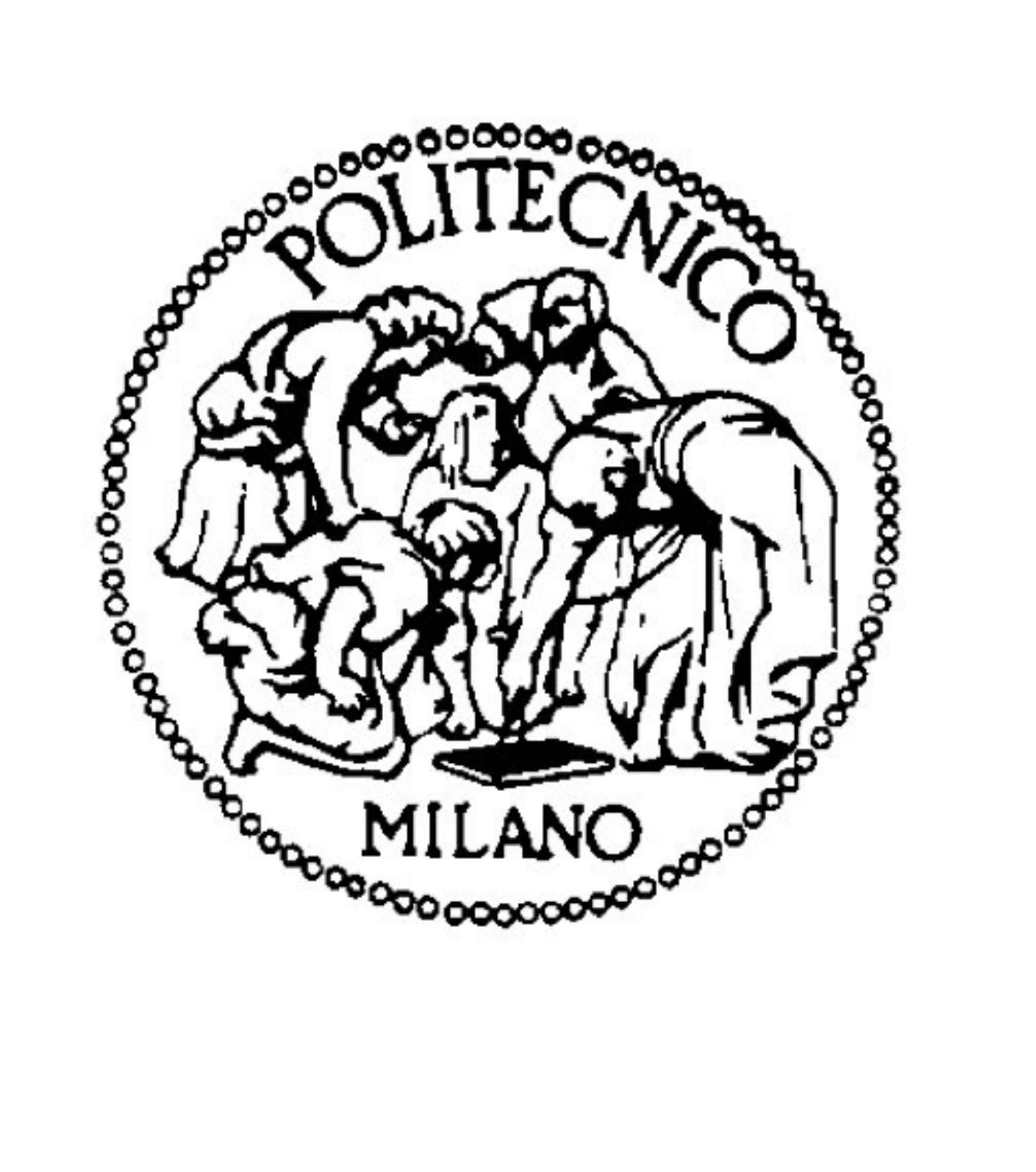}
\end{figure}
{\large \textbf{POLO TERRITORIALE DI COMO}}\\
{\small \textbf{Corso di Laura Specialistica in Ingegneria Informatica}}\\[1.5cm]
\linespread{1.6}
{\huge \textbf{Mountain Peak Detection\\ in Online Social Media\\ }}\\[1.0cm] 
\linespread{1}
\begin{flushleft}
{\bfseries Relatore:} Prof. Marco Tagliasacchi\\
{\bfseries Correlatore:} Prof. Piero Fraternali\\[1.5cm]
\end{flushleft}
{\bfseries Tesi di laurea di:}\\
Roman Fedorov\\
matr. 782110\\
\begin{flushright}
{\bfseries Anno Accademico} 2012/13
\end{flushright}
\thispagestyle{empty} \normalfont \cleardoublepage
\newpage
\chapter*{Abstract}


We present a system for the classification of mountain
panoramas from user-generated photographs followed by
identification and extraction of mountain peaks from those
panoramas. We have developed an automatic technique that,
given as input a geo-tagged photograph, estimates its FOV
(Field Of View) and the direction of the camera using a
matching algorithm on the photograph edge maps and a rendered
view of the mountain silhouettes that should be seen
from the observer's point of view. The extraction algorithm
then identifies the mountain peaks present in the photograph
and their profiles. We discuss possible applications in social
fields such as photograph peak tagging on social portals,
augmented reality on mobile devices when viewing a mountain
panorama, and generation of collective intelligence systems
(such as environmental models) from massive social
media collections (e.g. snow water availability maps based
on mountain peak states extracted from photograph hosting
services).
\thispagestyle{empty} \vspace*{.75truecm} \cleardoublepage
\newpage
\chapter*{Sommario}


Proponiamo un sistema per la classificazione di panorami montani e per l'identificazione delle vette presenti in fotografie scattate dagli utenti.
Abbiamo sviluppato una tecnica automatica che, data come input la foto e la sua geolocalizzazione, stima il FOV (Field of View o Angolo di Campo) e l'orientamento della fotocamera. Questo avviene tramite l'applicazione di un algoritmo di matching tra la mappa degli edge (bordi) della fotografia e alle silhouette delle montagne che dovrebbero essere visibili dall'osservatore a quelle coordinate.
L'algoritmo di estrazione identifica poi i picchi delle montagne presenti nella fotografia e il loro profilo. 
Verranno discusse alcune possibile applicazioni in ambito sociale come ad esempio: 
l'identificazione e tagging (marcatura) delle fotografie sui social network, 
realtà aumentata su dispositivi mobile durante la visione di panorami montani e
la generazione di sistemi di intelligenza collettiva (come modelli ambientali) dalle enormi collezioni multimediali dei social network (p.es. mappe della disponibilità di neve e acqua sulle vette delle montagne, estratte da servizi di condivisione di immagini).
\thispagestyle{empty} \vspace*{.75truecm} \cleardoublepage
\newpage
\chapter*{Acknowledgements}


Foremost, I would like to thank Prof. Marco Tagliasacchi and Prof. Piero Fraternali for the opportunity to work with them on a such interesting and exciting project, and the continuous support during the preparation of this thesis.

I would like to thank also:
\begin{itemize}
\item
Dr. Danny Chrastina (\url{http://www.chrastina.net}) for helping with the preparation of this document.
\item
Dr. Ulrich Deuschle (\url{http://www.udeuschle.de}) for his kind permission of using his mountain panorama generating web tool.
\item
Gregor Brdnik (\url{http://www.digicamdb.com}) for providing the database of digital cameras and their sensor sizes.
\item
Miroslav Sabo (\url{http://www.mirosabo.com}) for the photograph of the Matterhorn used in this thesis.
\item
All my friends and colleagues (who are too many to be listed) who have helped and contributed to this work.
\end{itemize}

Lastly, but most importantly, I would like to thank my parents (Alexey and Irina) for making it possible for me to get here, and Valentina for constant support and motivation.

\thispagestyle{empty} \vspace*{.75truecm} \normalfont \cleardoublepage
\tableofcontents
\thispagestyle{empty} \vspace*{.75truecm} \normalfont \cleardoublepage
\listoffigures
\thispagestyle{empty} \vspace*{.75truecm} \normalfont \cleardoublepage
\listoftables
\thispagestyle{empty} \vspace*{.75truecm} \normalfont \cleardoublepage

\pagestyle{plain}\renewcommand{\chaptermark}[1]{\markboth{\chaptername\ \thechapter.\ #1}{}} 
\renewcommand{\sectionmark}[1]{\markright{\thesection.\ #1}}         
\fancyhead[LE,RO]{\bfseries\thepage}    

\pagenumbering{arabic}
                                        
\fancyhead[RE]{\bfseries\leftmark}    
\fancyhead[LO]{\bfseries\rightmark}     
\renewcommand{\headrulewidth}{0.3pt} 

\chapter{Introduction}
\label{Introduction}
\thispagestyle{empty}

\vspace{0.5cm}

\noindent
The most suitable paradigm for representing this work is probably passive crowdsourcing, a field that is not trivial to define and that is not even a subcategory of crowdsourcing in the strict sense, as its name can suggest. A possible definition of the passive crowdsourcing discipline can be the union of crowdsourcing, data mining and collective intelligence (computer science fields that have much in common but are slightly different). These differences are often hard to notice due to the youth of these concepts and so the presence of a lot of confusion. For this reason the purpose of this chapter is to define these concepts unambiguously and to define the problem statement of this work.

\section{Human Computation}
The idea of the computer computation goal has always been that which Alan Turing expressed in 1950:
\begin{quotation}
{
\noindent{\emph{``
The idea behind digital computers may be explained by saying that these machines are intended to carry out any operations which could be done by a human computer.
''}
 Alan Turing \cite{Turing:1995:CMI:216408.216410}
}
}
\end{quotation}
Though current progress in computer science brings automated solutions of more and more complex problems, this idea of computer systems able to solve any problem that humans can solve is however far from reality. There are still a lot of tasks that cannot be performed by computers, and those that could be but are preferred to be computed by humans for quality, time, and cost reasons. These tasks lead to the field of human computation: a field that is hard to give a definition to, in fact several definitions of the term can be found: the most general, modern and suitable for our needs of which is probably that extracted from von Ahn's dissertation:
\begin{quotation}
{
\noindent{\emph{``
... a paradigm for utilizing human processing power to solve 
problems that computers cannot yet solve.
''}
 Luis von Ahn  \cite{VonAhn:2005:HC:1168246}
}
}
\end{quotation}
Human computation can be thought of as several approaches varying with the tasks involved, the type of persons involved in task completion, the incentive techniques used, and what type of effort the persons are required to make. It must be said that the classification of human computation is not an ordinary hierarchy with parents and children, but is instead a set of related concepts not necessary including one another. So a possible taxonomy of human computation (seen as a list of related ideas) has been produced by combining definitions given by Quinn et al. \cite{Quinn:2011:HCS:1978942.1979148} and Fraternali et al. \cite{Fraternali:2012:PHL:2263310.2263839} can be:
\begin{itemize}
\item
\emph{Crowdsourcing}:
this approach manages the distributed assignment of tasks to an open, undefined and generally large group of executors. The task to be performed by the executors is split into a large number of microtasks (by the work provider or the crowdsourcing system itself) and each microtask is assigned by the system to a work performer, who executes it (usually for a reward of a small amount of money). The crowdsourcing application (defined usually by two interfaces: for the work providers and the work performers) manages the work life cycle: performer assignment, time and price
negotiation, result submission and verification, and payment. In addition to the web interface, some platforms offer Application Programming Interfaces (APIs),whereby third parties can integrate the distributed work management functionality into their custom applications. Examples of crowdsourcing solutions are Amazon Mechanical Turk and Microtask.com \cite{Fraternali:2012:PHL:2263310.2263839}.
\item
\emph{Games with a Purpose (GWAPs)}:
these are a sort of crowsourcing application but with a fundamental difference in user incentive technique: the process of resolving a task is implemented as a game with an enjoyable user experience. Instead of monetary earning, the user motivation in this approach is the gratification of the playing process. GWAPs, and more generally useful applications where the user solves perceptive or cognitive problems without knowing, address task such as adding descriptive tags and recognising objects in images and checking the output of Optical Character Recognition (OCR) for correctness. \cite{Fraternali:2012:PHL:2263310.2263839}.
\item
\emph{Social Computing}:
a broad scope concept that includes applications and services that facilitate collective action and social interaction online with rich exchange of multimedia information and evolution of aggregate knowledge \cite{parameswaran2007social}. Instead of crowdsourcing, the purpose is usually not to perform a task. The key distinction between human computation and social computing is that social computing facilitates relatively natural human behavior that happens to be mediated by technology, whereas participation in a human computation is directed primarily by the human computation system \cite{Quinn:2011:HCS:1978942.1979148}.
\item
\emph{Collective Intelligence}: if seen as a process, the term can be defined as groups of individuals doing things collectively that seem intelligent \cite{malone-harnessing}. If it is seen instead as the process result, means the knowledge of any kind that is generated (even non consciously and not in explicit form) by the collective intelligence process. Quinn et al. \cite{Quinn:2011:HCS:1978942.1979148} classifies it as the superset of social computing and crowdsourcing, because both are defined in terms of social behavior. The key distinctions between collective intelligence and human computation are the same as with crowdsourcing, but with the additional distinction that collective intelligence applies only when the process depends on a group of participants. It is conceivable that there could be a human computation system with computations performed by a single worker in isolation. This is why part of human computation protrudes outside collective intelligence \cite{Quinn:2011:HCS:1978942.1979148}.
\item
\emph{Data Mining}:
this can be defined broadly as the application of specific algorithms for extracting patterns from data \cite{Fayyad96knowledgediscovery}. Speaking about human-created data the approach can be seen as extracting the knowledge from a certain result of a collective intelligence process. Creating this knowledge usually is not the goal of the persons that generate it, in fact often they are completely unaware of it (just think that almost everybody contributes to the knowledge of what are the most popular web sites just by visiting them: they open a web site because they need it, not to add a vote to its popularity). Though it is a very important concept in the field of collective intelligence, machine intelligence applied to social science and passive crowdsourcing (that will be defined in the next section) is a fully automated process by definition, so it is excluded from the area of human computation.
\item
\emph{Social Mobilization}:
this approach deals with social computation problems where the timing and the efficiency is crucial. Examples of this area are safety critical sectors like civil protection and disease control.
\item
\emph{Human Sensors}:
exploiting the fact that the mobile devices tend to incorporate more and more sensors, this approach deals with a real-time collection of data (of various natures) treating persons with mobile devices as sensors for the data. Examples of these applications are earthquake and other natural disaster monitoring, traffic condition control and pollution monitoring.
\end{itemize}

\section{Passive Crowdsourcing}
\begin{figure}[h!]
	\centering
	\includegraphics[width=0.85\columnwidth]{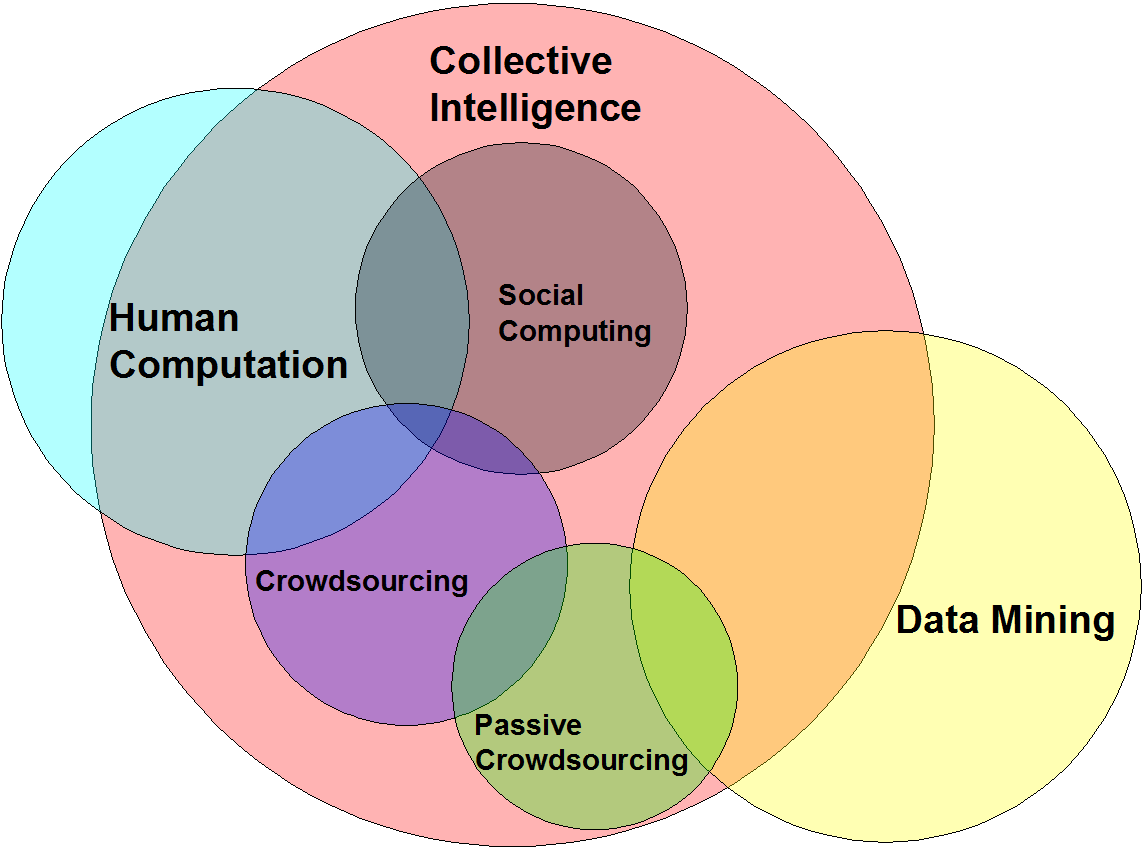}
	\caption{Proposed taxonomy of human computation including passive crowdsourcing.}
	\label{fig:crowdsourcingTaxonomy}
\end{figure}

The goal of any passive human computation system is to exploit the collective effort of a large group of people to retrieve the collective intelligence this effort generates or other implicit knowledge. In this study case the collective effort is taking geo-tagged photographs of mountains and publishing them on the Web, the intelligence deriving from this collection of photographs is the availability of the appearances of a mountain through time, the knowledge we want to extract having these visual appearances of mountains is the evolution of its environmental properties in time (which can be for example snow or grass presence at a certain altitude) or using some ground truth data even to predict these features where the use of physical sensors for those measurements is difficult or impossible (i.e. snow level prediction at high altitudes). Passive crowdsourcing is an approach that is not trivial to classify within the described taxonomy: it can be best classified in an area that includes: \emph{crowdsourcing} for the fact of exploiting the effort of human computation, \emph{collective intelligence} since its extraction is the primary goal of the approach and \emph{data mining} as it refers to the procedure of extracting some results of human computation from the public Web data. Figure \ref{fig:crowdsourcingTaxonomy} shows the taxonomy proposed by Quinn et al. \cite{Quinn:2011:HCS:1978942.1979148} with this proposal of passive crowdsourcing collocation.

The main advantage of passive crowdsourcing with respect to traditional crowdsourcing is the enormous availability of data to collect and very reduced costs of its retrieval, a significant disadvantage, on the other hand, is the form of the data: not always perfectly suitable for the goals and almost always needing to be processed and analyzed before being used. In other words, traditional crowdsourcing asks the user to shape the data in the input in a certain way, paying for this; passive crowdsourcing instead retrieves the result of past work, but shaped as it is, so processing it next. Just think about a study of the public opinion trends of a certain commercial product, two different ways to collecting these opinions can be used:
\begin{itemize}
\item
\emph{Crowdsourcing}: a certain amount of money is paid to each person who gives his proper opinion about the product being studied (approach also called crowdvoting). This method guarantees perfectly shaped collected data: the form of the questions and answers can be decided and modeled easily, but obviously the availability of the data to be collected will be limited by the number of the people ready to take that survey, and the costs of collecing a huge dataset of opinions will not surely be low.
\item
\emph{Passive Crowdsourcing}: the publicly available web data is full of opinions of the customers of a certain product (think about a customer that posts a photograph of that product, or comments a photograph uploaded by someone else, declaring their personal opinion about that product), the cost of retrieving these photographs and comments are almost null and the availability of this content is enormous. The big problem anyway is the shape of this data: given a photograph the algorithm must decide whether it is pertinent to the study or not (if the object in the photograph is the product being looked for) and to estimate if the opinion expressed by the user is positive or negative.
\end{itemize}

In the context of this work the huge amount of available data is fundamental, so the passive crowdsourcing approach is prefered, and the purpose of this work is exactly to deal with the problem of data shaping and analyzing.

Another significant advantage of passive crowdsourcing (and very important for this work) is the availability of the implicit information the user himself is unaware of: if a person is asked whether the peak of the Matterhorn (a mountain in the Italian Alps) was covered with snow or not in August of 2010 - he does not remember, but an appropriate image processing technique can extract this information from the photograph the user posted on his social network profile during his vacation.

\section{User Generated Content}
The amount of available user generated media content on the Web nowadays is reaching unprecedented mass: Facebook alone hosts 240 billion photographs and gets 300 million new ones every day \cite{Doherty2010Lifelog}. This massive input allows collective intelligence applications to reach unprecedented results in innumerable domains, from smart city scenarios to land and environmental protection.

A significant portion of this public dataset are geo-tagged photographs. The availability of geo-tags in photos results from two current trends: first, the widespread habit of using the smartphone as a photo camera; second, the increasing number of digital cameras with an integrated GPS locator and Wi-Fi module. The impact of these two factors is that more and more personal photographs are being published on social portals and more and more of them are precisely geo-tagged \cite{npdPhotosSmartphones}. A time-stamped and geo-located photo can be regarded as the state of one or more objects at a certain time. From a collection of photographs of the same object at different moments, one can build a model of the object, study its behavior and evolution in time, and build predictive models of properties of interest.

However, a problem in the implementation of collective intelligence applications from visual user generated content (UGC) is the identification of the objects of interest: each object may change its shape, position and appearance in the UGC, making its tracking in a set of billions of available photographs an extremely difficult task. In this work we aim at realizing a starting point for the applications that generate collective intelligence models from user-generated media content based on object extraction and model construction in a specific environmental sector: the study of mountain conditions. We harness the
collective effort of people taking pictures of mountains from different positions and at different times of the year to produce models describing the states of selected mountains and their changing snow conditions over time. To achieve this objective, we need to address the object identification problem, which for mountains is more tractable than the general case described above, thanks to the fact that mountains are among the most motionless and immutable objects present on the planet. This problem of mountain identification in fact will be the goal of this work.

\section{Problem Statement}
Given a precisely geo-tagged photograph, the goal of this work is to determine whether the photograph contains a mountain panorama, if yes, estimate the direction of view of the photo camera during the shot and identify the visible mountain peaks on the photograph.

We will describe in the detail the proposed algorithm, how it has been implemented and tested with the result of successfully matching 64.2\% of the input geo-tagged photographs.

The algorithm of peak detection that is  presented can be used in applications where the purpose is to identify and tag mountains in photographs provided by users. The algorithm can also be used for the creation of mountain models and of their related properties, such as, for example, the presence of snow at a given altitude.

Two representative examples of the first type of usage are:
\begin{itemize}
\item Mountain peak tagging of user-uploaded photographs on photo sharing platforms that allow anyone browsing that photograph to view peak names of personal interest.
\item Augmented reality on mobile devices with real-time peak annotation on the device screen while in camera mode.
\end{itemize}

An example of the usage for environmental model building is the construction of a model for correcting ground and satellite based estimates of the Snow Water Equivalent (SWE) on mountains peaks (which is described in the Conclusions and Future Work section).

\section{Document Structure}
In the next chapter several past works will be discussed, each one relevant in its own way and field to this work: from passive crowdsourcing and influenza surveillance to image processing and vision-based UAV navigation.

In the third chapter the proposed algorithm itself is described in detail and an efficient vector-based matching algorithm is explained.

In chapter four the realized implementation of the discussed algorithm is explained, including all improvement techniques developed (even those that have been rejected after the validation phase).

In the fifth chapter the results of the tests performed on the implemented algorithm are listed with the data set structure and error metric description.

Finally in the last chapter the conclusions about this work are drawn with the possible future direction of this project.

\chapter{Related Work and State of the Art}
\label{Related Work and State of the Art}
\thispagestyle{empty}

\vspace{0.5cm}

\noindent
This work combines many disciplines of social computing, image processing, machine intelligence and environmental modeling with the relative emblematic problems: passive crowdsourcing, object identification (in particular mountain boundaries) and pose estimation, collective intelligence extraction and knowledge modeling, snow level estimation. All of the these problems have been largely analyzed and studied recently, but only a very small part of them combines several of the listed problems.

Quoting all the works in those fields would be impossible, so in this chapter several examples of works from each of these fields (often combined together) will be discussed, from social problems and those closely inherent to this work, to examples of non-social applications, emphasizing in fact the wide possibility of goals that can be reached with these approaches.

The fundamental concept of computational social science is treated by Lazer et at. \cite{lazer2009life}, explaining the trend of social science, moving from the analysis of individuals to the study of society: nowadays almost any action performed, from checking email, making phone calls and checking our social network profile to going for a walk, driving to the office by car, booking a medical check-up or even paying at the supermarket with a credit card leaves a digital fingerprint. The enormous amount of these fingerprints generates data, that pulled together properly give an incredibly detailed picture of our lives and our societies. Understanding these trends of societal evolution and so also individuals changing is a complex task, but with an unprecedented amount of potential and latent knowledge waiting to be extracted. The authors discuss the problems of this approach such as managing of privacy issues (think about the NRC report on GIS data \cite{national2007Putting} describing the possibility to extract the individual information even from well anonymized data, or the online health databases which were pulled down after a study revealed the possibility of confirming the identities \cite{dnaBlocked}). The lack of the both approach and infrastructure standards in this emerging field is also discussed:
\begin{quotation}
{
\noindent{\emph{``
The resources available in the social sciences are significantly smaller, and even the physical (and administrative) distance between social science departments and engineering or computer science departments tends to be greater than for the other sciences. The availability of easy-to-use programs and techniques would greatly magnify the presence of a computational social science. Just as mass-market CAD software revolutionized the engineering world decades ago, common computational social science analysis tools and the sharing of data will lead to significant advances. The development of these tools can, in part, piggyback on those developed in biology, physics and other fields, but also requires substantial investments in applications customized to social science needs.
''}
 Lazer et at. \cite{lazer2009life}
}
}
\end{quotation}

\section{Passive Crowdsourcing and Collective Intelligence}
An important work in this field (even if the authors do not use the term of passive crowdsourcing, but it is exactly the type of problem solving we mean this term for) is performed by Jin et al. \cite{Jin:2010:WSM:1873951.1874196} in their study of society trends from photograph analysis. The authors propose a method for collecting information to identify current social trends, and also for the prediction of trends by analyzing the sharing patterns of uploaded and downloaded social multimedia. Each time an image or video is uploaded or viewed, it constitutes an implicit vote for (or against) the subject of the image. This vote carries along with it a rich set of associated data including time and (often) location information. By aggregating such votes across millions of Internet users, the authors reveal the wisdom that is embedded in social multimedia sites for social science applications such as politics, economics, and marketing \cite{Jin:2010:WSM:1873951.1874196}.

Given a query, the relevant photographs with relative metadata are extracted from Flickr, and the global social trends are estimated. The motivation for introducing this approach is the low cost of crawling this information for the companies and the industries, as well as the possibility to analyze this data almost instantaneously with respect to common surveys. The implementation of the proposal is also discussed, with several tests in various fields that gave incredibly promising results:
\begin{itemize}
\item
\emph{Politics}: the popularity scores analysis of the candidates Obama and McCain during the USA president elections of 2008 gave the resulting trends which were correct within a tenth of a percent of the real election data.
\item
\emph{Economics}: a product distribution map (with \emph{iPod} as the example) around the world over time was successfully drawn.
\item
\emph{Marketing}: the sales of past years of several products such as music players, computers and cellular phones were estimated and the results actually match the official sales trends of those products.
\end{itemize}

Although being an innovative and efficient approach, it does not deal with the visual content of the images themselves (the authors also highlight this fact, declaring the intent to add this analysis in the future). An example of collective intelligence extraction using image content is the work of Cao et al. \cite{conf/icassp/CaoLGJHH10} that presents a worldwide tourism recommendation system. It is based on a large-scale geotagged web photograph collection and aims to suggest with minimal input to tourists the destination they would enjoy. By taking more than one million of geotagged photographs, dividing them into clusters by geographical position and extracting the most representative photographs for each area, the system is able to propose destinations to the user having in input the set of keywords and images of the places the user likes. From a conceptual point of view this system tries to simulate a friend that knows your travel tastes and suggests destinations for your new journey, but with the difference that this virtual friend has visited millions of places all around the world. This lies exactly in the concept of collective intelligence: exploiting the effort of hundreds of thousands of people uploading their travel photographs, the authors build a knowledge on what the various world tourism destinations feel like.

\subsection{Real-Time Social Monitoring}
An approach related to the concept of Human Sensors treated in the previous chapter is the process of monitoring the online activity of the persons to predict in advance (or at least to identify quickly) the occurrence of some phenomena. The activity to be monitored and the phenomena to detect can be very different. The most popular online activity to monitor is for sure the web searches of the users, such as in works of Polgreen et al. \cite{Polgreen_usinginternet}, Ginsberg et al. \cite{citeulike:3681665} and Johnson et al. \cite{15361003} (in Johnson et al. also the access logs to health websites are analyzed) that propose influenza surveillance exploiting web search query statistics, examining the connection between searches for influenza and actual influenza occurrence and finding strong relationships. Another example of web query monitoring is the prediction of macroeconomic statistics by Ettredge et al. \cite{Ettredge:2005:UWS:1096000.1096010}, in particular the prediction of unemployment rate trend based on the frequency of search terms likely used by people seeking employment.

Another important source of activity are social networks, for example Twitter messages, used by Culotta \cite{Culotta:2010:TDI:1964858.1964874} in the monitoring of influenza, similar to the references described above, and by Sakaki et al. \cite{Sakaki:2010:EST:1772690.1772777} for detecting the earthquakes.

\section{Object Identification and Pose Estimation}
The pose estimation of the photograph will be the key problem of this work, even if the estimated variable is only the orientation and not the position (that is given in input), and is a very commonly treated problem. Several examples will be listed here with problems, each one with its own elements in common with this work.

An example of a relatively different problem of pose estimation with respect to this work is the estimation of the geographic position of a photograph proposed by Hays and Efros \cite{Hays:2008:im2gps}: though the estimation of the position is performed with the analysis of the visual content of the image, a purely data-driven scene matching approach is applied to estimate a geographic area the photograph belongs to.

Ramalingam et al. \cite{RBSB09} instead present a work that at first sight can seem the opposite of this one (instead of estimating the orientation given the position they estimate the position given the orientation: always perpendicular to the terrain) but it is very similar: the authors describe a method to accurately estimate the global position of a moving car using an omnidirectional camera and untextured 3D city models. The idea of the algorithm is the same: estimate the pose by matching the input image to a 3D model (city model in this case, elevation model in case of our work). The described algorithm extracts the skyline from an omni-directional photograph, generates the virtual fisheye skyline views and matches the photograph a to view, estimating in this way the position of the camera.

Other works about pose estimation given 3D models of cities and buildings have been recently published, such as world wide pose estimation using 3D point clouds by Li et al. \cite{Li:2012:WPE:2402940.2402943} in which the SIFT features are located and extracted in the photograph and matched with the worldwide models. The particular point of this pose estimation is that it does not use any geographical information, but it estimates the position, orientation and the focal length (and so the field of view).

Baatz et al. \cite{Baatz:2012:LCM:2125160.2125170} addresses the problem of place-of-interest recognition in urban scenarios exploiting 3D building information, giving in output the camera pose in real world coordinates ready for augmenting the cell phone image with virtual 3D information. Sattler et al. instead deals with the problem of the 2D-to-3D correspondence computation required for these cases of pose estimation of urban scenes, demonstrating that direct 2D-to-3D matching methods have a considerable potential for improving registration performance.

An innovative idea of photograph geolocalization by learning the relationship between ground level appearance and overhead appearance and land cover attributes from sparsely available geotagged ground-level images \cite{LinCrossView} was introduced by Lin et al. \cite{LinCrossView}: unlike traditional geolocalization techniques it allows the geographical position of an isolated photograph to be identified (with no other geotagged photographs available in the same region). The authors exploit two previously unused data sets: overhead appearance and land cover survey data. Ground and aerial images are represented using HoG \cite{Dalal:2005:HOG:1068507.1069007}, self-similarity \cite{shechtman2007matching}, gist \cite{Oliva:2001:MSS:598425.598462} and color histograms features. For each of these data sets the relationship between ground level views and the photograph data is learned and the position is estimated by two proposed algorithms that are also compared with three other pre-existing techniques:
\begin{itemize}
\item
\emph{im2gps}:
proposed by Hays and Efros \cite{Hays:2008:im2gps} already described a few lines above, that does not make use of aerial and attribute information and can only geolocate query images in locations with ground-level training imagery.
\item
\emph{Direct Match (DM)}:
matches the same features for ground level images to aerial images with no translation, assuming that the ground level appearance and overhead appearance are correlated.
\item
\emph{Kernelized Canonical Correlation Analysis (KCCA)}:
is a tool to learn the basis along the direction where features in different views are maximally correlated, used as a matching score. It however presents significant disadvantages: singular value decomposition for a non-sparse kernel matrix is need to solve the eigenvalue problem, making the process unfeasible as training data increases, and secondly, KCCA assumes one-to-one correspondence between two views (in contrast with the geolocalization problem where it is common to have multiple ground-level images taken at the same location) \cite{Hays:2008:im2gps}. 
\end{itemize}

The methods proposed by the authors are instead:
\begin{itemize}
\item
\emph{Data-driven Feature Averaging (AVG)}:
based on the idea that well matched ground-level photographs will tend to have also similar aerial and land cover attributes, this technique translates the ground level to aerial and attribute features by averaging the features of good scene matches.
\item
\emph{Discriminative Translation (DT)}:
an approach that extends AVG with also a set of negative training samples, based on the intuition that the scenes with very different ground level appearance will have distinct overhead appearance and ground cover attributes (assumption obviously hypothetical and not always true).
\end{itemize}

In the performed tests the algorithm was able to correctly geolocate 17\% of the isolated query images, compared to 0\% for existing methods.

\subsection{Mountain Identification}
All the pose estimation works described so far in this chapter deal with urban 3D models, as does the majority of pose estimation research. This is not surprising since an accurate 3D model is fundamental for this kind of task, and the massive increase of the 3D data of buildings and cities in the last few years makes these studies possible. Apart from urban models however, there is another type of 3D data that is largely available, which evolved much earlier than urban data (even if usually with lower resolution): terrain elevation data. The elevation data, presented usually as a geographical grid with the altitude for each point, can be easily seen as a 3D model, and the most interesting objects formed by these models are for sure mountains. For this reason also mountains are sometimes the identified objects in the pose estimation task, here several examples of these works will be discussed.

The most significant work in this sector is probably that presented by Baboud et al. \cite{Baboud2011Alignment}, which given an input geotagged photograph, introduces the matching algorithm for correct overlap identification between the photograph boundaries and those of the virtually generated panorama (based on elevation datasets) that should be seen by the observer placed in the geographical point where the photograph has been taken from. This algorithm, which will be discussed in detail further, is the starting point of this thesis work. It must be highlighted however, that the goal of the authors is peak identification for the implementation of photograph and video augmentation; this work instead aims to identify mountain peaks to extract the appearances of the mountain for environmental model generation.

Another important work, dealing not only with the direction, but also with the position estimation of a mountain image, was written by Naval et al. \cite{Jr97estimatingcamera}. Their algorithm does not work with the complete edges of the image but only with the skyline extracted using a neural network. The position and the orientation is then computed by nonlinear least squares.

A proposal that exploits the skyline and mountain alignment is the vision-based UAV navigation described by Woo et al. \cite{WooSLKK07}, that with pose estimation in a mountain area (a problem similar to the other described works) introduces the possibility of vision-based UAV navigation in a mountain area using an IR (Infra-Red) camera by identifying the mountain peaks. This is a navigation method that is usually performed by extracting features such as buildings and roads, not always visible and available, so mountain peak and skyline recognition brings big advantages.

A challenging task with excellent results is described by Baatz et al. \cite{Baatz:2012:LSV:2403006.2403045} with a proposal of an algorithm that given a photograph, estimates its position and orientation on large scale elevation models (in case of the article a region of 40000~km$^2$ was used, but in theory the technique can be applied to position estimation on a world scale. The algorithm exploits the shape information across the skyline and searches for similarly shaped configurations in the large scale database. The main contributions are a novel method for robust contour
encoding as well as two different voting schemes to solve the large scale camera pose recognition from contours. The first scheme operates only in descriptor space (it checks where in the model a panoramic skyline is most likely to contain the current query picture) while the second scheme is a combined vote in descriptor and rotation space \cite{Baatz:2012:LSV:2403006.2403045}. The original six-dimensional search is simplified by the assumptions that the photograph has been taken close to the terrain level and the photograph has usually only a small roll with respect to the horizon (both assumptions were made also during the development of this work algorithm). Instead of supposing to have the right shot position, this technique renders the terrain view from a digital elevation model on a grid defined by distances of approximately 100~m~$\times$~100~m for a total number of 3.5 million cubemaps (this is how the authors call the renders). The key problem of the work is in fact, the large scale search method to efficiently match the query skyline to one of the cubemaps. Given a query image, sky/ground segmentation is performed following an approach based on unary data costs \cite{Martin:2004:LDN:977249.977379,Luo:2002:PAD:839290.842704} for a pixel being assigned sky or ground. The horizon skyline then is represented by a collection of vector-quantized local contourlets (contour words, similar in spirit to visual words obtained from quantized image patch descriptors) that are matched to the collected cubemaps with a voting stage that retrieves the most probable cubemaps to contain the query skyline, the top 1000 candidates are then analyzed with geometric verification using iterative closest points to determine a full 3D rotation. Evaluation was performed on a data set of photographs with manually verified GPS tags or given location, and in the best implementation 88\% of the photographs were localized correctly.

\section{Environmental Study}
In this section the problem of snow level and snow water equivalent (SWE) estimation will be described, with the current state of the art and used methods as well as the benefits that an environmental model based on passive crowdsourcing photograph analysis can bring.

Snow Water Equivalent (SWE) is a common snowpack measurement. It is the amount of water contained within the snowpack. It can be thought of as the depth of water that would theoretically result if you melted the entire snowpack instantaneously. To determine snow depth from SWE you need to know the density of the snow. The density of new snow ranges from about 5\% when the air temperature is 14\degree F, to about 20\% when the temperature is 32\degree F. After the snow falls its density increases due to gravitational settling, wind packing, melting and recrystallization \cite{whatIsSWE}. It is a very important parameter for industry and agriculture, and its correct estimation is one of the main problems facing the environmental agencies.

These measurements are usually performed with physical sensors and stations that are very sparse, so the need of a map of snow and SWE distribution introduces a key problem: the interpolation of this data. Interpolation is usually done (due to the lack of the sophisticated physical models dealing with altitude and temperature and the absence of other supporting data) with relatively rough methods such as:
\begin{itemize}
\item
linear or almost linear interpolation of the data, lacking of precision due to the low spatial density of the measurement stations and the physical laws and phenomena that change the snow level and density radically with changes in altitude
\item
combining the interpolated data with the satellite ground images, removing the snow estimation from the areas where the satellite image indicates its absence: although being a first step in image content analysis for the refining of snow data, it indicates only a binary presence or absence of the snow and has anyway a low spatial density due to its low resolution.
\end{itemize}

The availability of the estimation of snow properties retrieved from mountain photograph analysis, and the past years snow measure ground truth in the areas and altitudes different from those that are measured by physical sensors, can bring significant supporting data for a better interpolation and more precise snow cover maps.

\subsection{Passive Crowdsourcing and Environmental Modeling}
An important step in environmental modeling exploiting passive crowdsourcing was made by Zhang et al. \cite{Zhang:2012:MPW:2187836.2187938} in a work that studies the problem of estimating geotemporal distributions of ecological phenomena using geo-tagged and time-stamped photographs from Flickr. The idea is to estimate a given ecological phenomena (the presence or absence of snow or vegetation in this case) on a given day at a given place, and generate the map of its distribution. A Bayesian probabilistic model was used, assuming that each photograph taken in the given time and place is an implicit vote to the presence or the absence of snow in that area. Exploiting machine learning, the probability that a photograph contain snow is estimated based both on the metadata of the photograph (tags) and the visual features (a simplified version of GIST classifier augmented with color features was used). The estimation of the results (which was made by the authors thanks to the fact that for the two phenomena studied, snowfall and vegetation cover, large-scale ground truth is available in the form of observations of satellites \cite{Zhang:2012:MPW:2187836.2187938}) brought very promising results: a daily snow classification for a 2 year period for four major metropolitan areas (NYC, Boston, Chicago and Philadelphia) generates results with precision, accuracy, recall and f-measure equal to approximately $0.93$.

Even if the algorithm proposed by authors uses a simple probabilistic model (presence or absence of snow and vegetation cover) it is an important introduction to the field of environmental and ecological phenomena analysis by mining photo-sharing sites for geo-temporal information about these phenomena. The goal of this work is to propose an image processing technique that brings these environmental studies to a new level.
\chapter{Proposed Approach}
\label{Proposed Approach}
\thispagestyle{empty}

\vspace{0.5cm}

\noindent
In this chapter we describe the proposed approach for the procedure of mountain identification, from the analysis of the properties of the photo camera used for the shot to the identification of the position in the photograph of each mountain peak.

The matching algorithm is partially based on an mountain contour matching technique proposed by Baboud et al. \cite{Baboud2011Alignment}, so first this matching algorithm will be discussed with particular emphasis on its advantages and disadvantages and then our proposal will be described in detail.

\section{Analysis of Babound et al.'s algorithm}
The basic idea and the purpose of the algorithm is the same as that of this work: given a geotagged photograph in input, identify the mountain peaks present in it, by exploiting the elevation data and the dataset of peaks with their positions.
The algorithm in question treats the input image as spherical, and searches for the rotation on $SO(3)$ that gives the right alignment between the input and another spherical image representing the mountain silhouettes viewed by the shot point of the input photograph in all directions. This image is generated by the 3D terrain model based on a digital elevation map. A vector cross-correlation technique is proposed to deal with the matching problem: the edges of the images are modeled as imaginary numbers, the candidates for the estimated direction of view are extracted, and a robust matching algorithm is used to identify the correct match.

As will follow from the next sections, several changes with respect to the original algorithm have been introduced, most relevant among them are:
\begin{itemize}
\item
Instead of dealing with the elevation models we rely on an external service, which generates the panorama view given the geographic coordinates and a huge set of parameters. The reason of this choice is the possibility to concentrate the work on matching technique, as well as the reliability and precision of a tool improved over the years.
\item
Instead of supposing the field of view to be known, its estimation given a photograph and the basic EXIF information will be discussed together with the photograph scaling problem, that allows successful matching by a non scale-invariant algorithm.
\item
Instead of searching for the orientation of the camera during the shot in three dimensions, we suppose that the observer's line of sight is always perpendicular to the horizon (no photograph in the test dataset had a significant tilt angle with respect to the horizon). This assumption simplifies the computational effort of the algorithm, by moving from considering the images as spherical to cylindrical.
\item
Once the camera direction is successfully estimated, the mountain peak alignment between the photograph and the panorama may still presents non-negligible errors due to inaccuracies of the estimated position of the observer. We will deal this problem, which is not treated in the original algorithm.
\end{itemize}

\section{Overview of proposed algorithm}
\begin{figure}[h!]
	\centering
	\includegraphics[width=0.65\columnwidth]{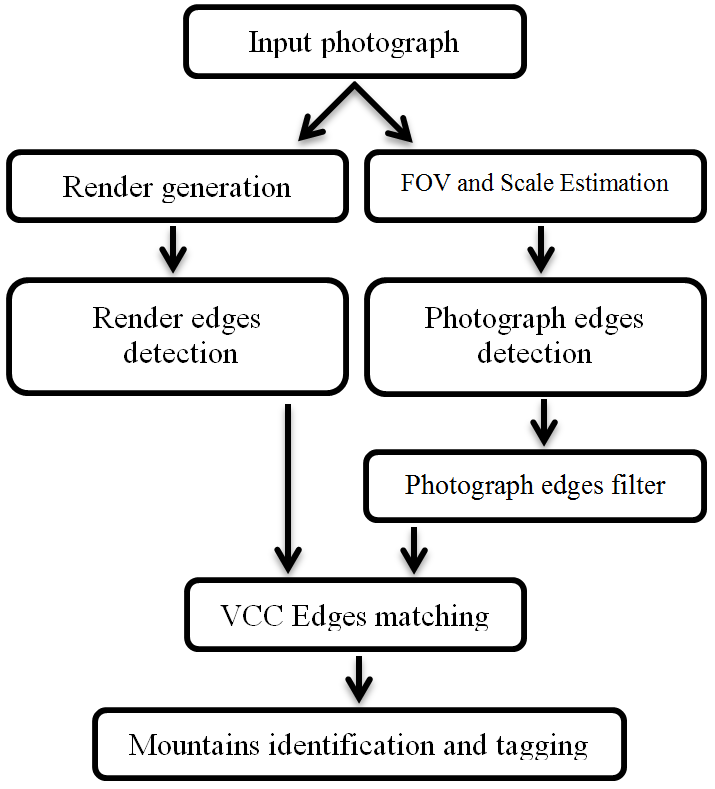}
	\caption{Schema of the mountain peak tagging algorithm.}
	\label{fig:algorithmSchema}
\end{figure}

The process of analyzing a single photograph containing a mountain panorama to identify individual mountain peaks present in it consists of several steps, shown in Figure \ref{fig:algorithmSchema}.

First the geotagged photograph is processed in order to evaluate its Field Of View and to scale it making the photograph match precisely the rendered panorama image, next a 360-degree rendition of the panorama visible at the location is generated from the digital elevation models.
After this the direction of the camera during the shot is estimated by extracting the edge maps of both photograph and rendered panorama images and matching them. Finally the single mountain peaks are tagged based on the camera angle estimation.

\section{Detailed description of the algorithm}

\subsection{Render Generation}
The key idea of the camera view direction estimations lies in generating a virtual panorama view of the mountains that should be seen by the observer from the point where the photograph was taken, and then matching the photograph to this panorama.
The generation of this virtual view is possible due to the availability of terrain elevation datasets that cover a certain geographical area with a sort of grid identifying the terrain elevation in each grid point. Depending on the source of the elevation data, the precision and spatial density of this grid can vary significantly, but usually it is very precise, reaching even a spatial density of 3 meters in public datasets (such as USGS, http://ned.usgs.gov).

Though the accuracy of elevation models is crucial for our purposes, since an exact render is the basis for a correct match, we are not necessary looking at extremely high-resolution datasets as the mountains we are going to generate are usually located at a significant distance from the observer (starting from few hundreds of meters to tens of kilometers). For this project the use of an external service that generated these rendered panoramas was preferred instead of creating them from scratch using the elevation data. The advantage of this choice is the possibility to avoid dealing directly with digital elevation maps and with optical and geometric calculations, in order to provide the panorama exactly as it should be seen by a human eye or photo camera lens. Such tools thus free the system from laborious calculations, allowing simply to choose the parameters needed for the generation of the panorama such as the observer's position, altitude and angle of gaze.

\begin{figure}[h!]
	\centering
	\includegraphics[width=1.0\columnwidth]{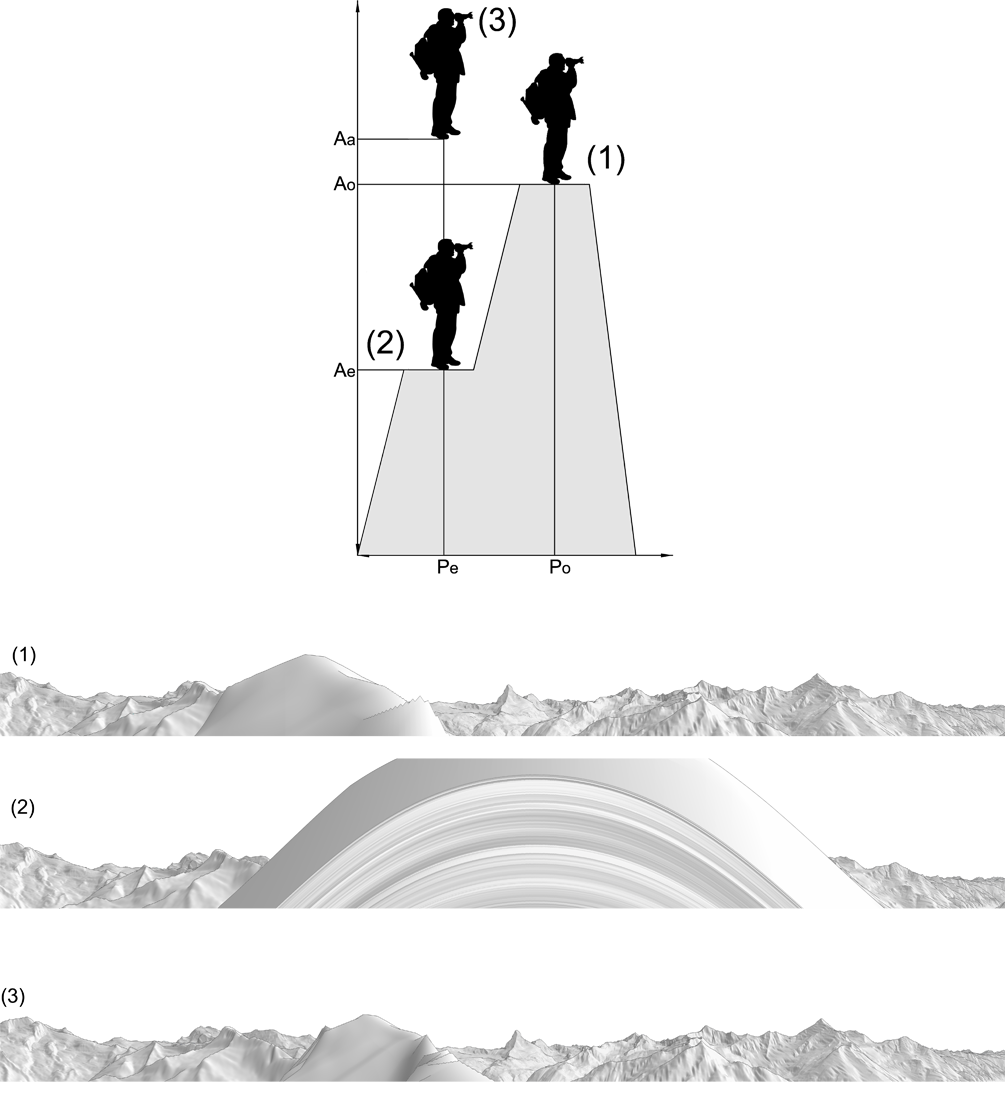}
	\caption{Example of wrong position estimation problems with view positions and the relative generated panoramas. 1 - Original viewer's position and altitude, 2  - estimated position and wrong altitude, 3 - elevated estimated position used for the final panorama generation.}
	\label{fig:mountainView}
\end{figure}

The choice of the observer's altitude deserves to be mentioned separately, as it is a critical point: clearly the ideal value that can be set (with an ideal elevation model) is the real altitude of the observer during the shot (which, after studying the available datasets, we consider more than legitimate to suppose to be on the terrain surface, so we estimate the altitude of the observer as the terrain altitude in that position). This information is readily available, but it brings some problems due to the uncertainty of the geotag of the photograph: let us imagine an observer standing on a peak or a ridge of a mountain - he has a broad view in front of him, but it is enough for him to take a few steps back and the panorama he was viewing becomes completely covered by the facade of the mountain in front of him. The same issue occurs with the photographs: a photograph taken from a peak or a ridge of a mountain (very frequently the case in public domain collections of mountain photographs). The error of estimating the photo camera position of few meters can lead to an overlay of a significant portion of the panorama. An intuitive technique that can walk around this problem is to add some constant positive offset to the estimated altitude to "raise" the observer above the terrain obstacles that have appeared due to the errors in position estimation. Figure \ref{fig:mountainView} shows in a simplified way the problem and its resolution with the corresponding generated panoramas.

\subsection{Field of View Estimation and Scaling}
Unless a scale invariant matching technique is going to be used (and it is not the case of this algorithm), once the input image and an image representing the expected panorama view are ready, the first problem is that in order to be correctly matched the objects contained in both images (in our case the objects to be matched are mountains) must have the same pixel size. Since we assume that both represent the view of an observer from the same geographical point, the same pixel dimension of the object involves also the same angular dimensions (the angle a certain object occupies in the observer's view). So we define our first photograph analysis problem as searching for the right scaling of the input photograph to have the same angular dimensions for the same mountains present in the photograph and the rendered panorama. We can write down the problem as finding a scale factor $k$ such that $$k\frac{s_{p}}{a_{s}} = \frac{s_{r}}{a_{r}}$$ where $s_{p}$ and $a_{p}$ are respectively the pixel size and the angular size of the input photograph and $s_{r}$ and $a_{r}$ are similarly the pixel size and the angular size of the rendered panorama.

We expect the panorama to be exact, so we consider the mountains to have the same width/height ratio both on the photograph and on the panorama, so the relationship described above can be equivalently applied both to the horizontal and vertical dimensions: we will work with the horizontal dimenstion because the angular width of the panorama does not need any calculation (it is always equal to the round angle, $2\pi$). Defining the Field Of View (FOV) of an image as its angular width we can rewrite the relationship as $$k*\frac{w_{p}}{FOV_{s}} = \frac{w_{r}}{FOV_{r}} = \frac{w_{r}}{2\pi}$$ where $w$ and $FOV$ stands for the pixel width and the FOV of respectively the photograph and the rendered panorama.

\begin{figure}[h!]
	\includegraphics[width=1.0\columnwidth]{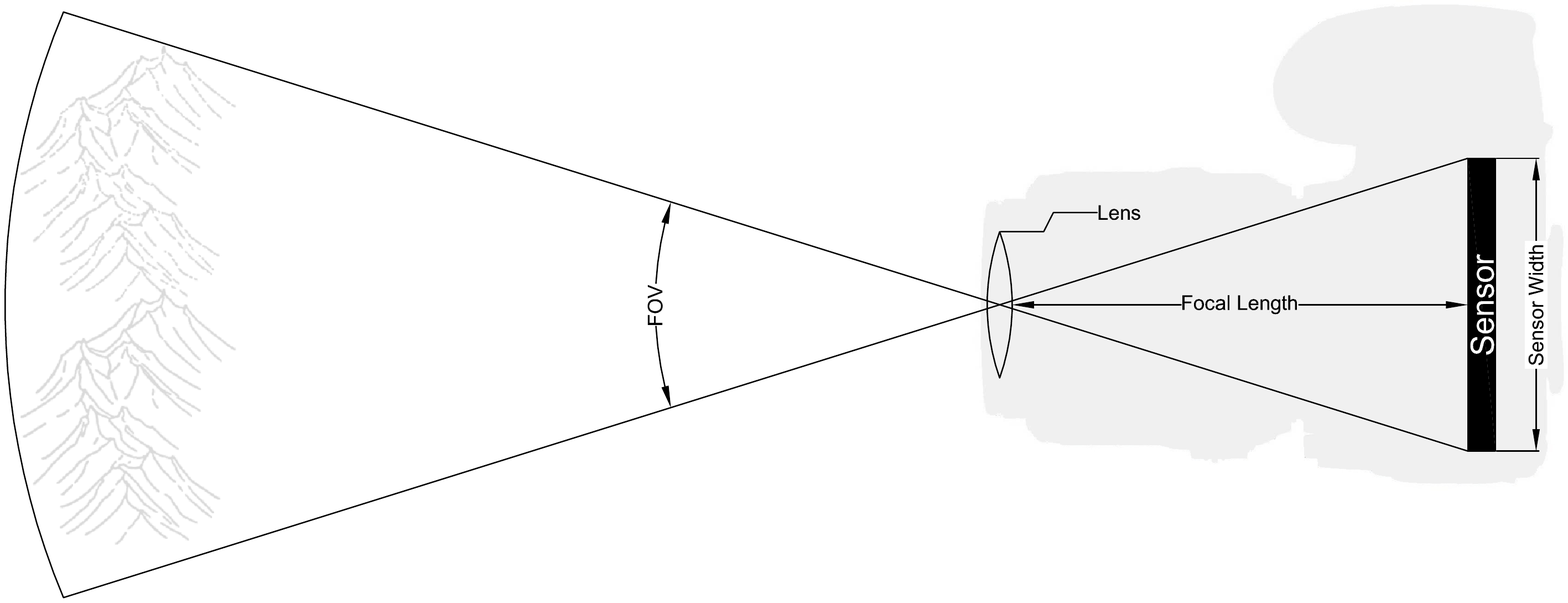}
	\caption{A simplified schema of digital photo camera functioning.}
	\label{fig:fovSchema}
\end{figure}

Before explaining how to estimate the FOV of the input photograph we must introduce some brief concepts regarding digital photo camera structure and optics laws: in a very simplified way a photo camera can be seen as a lens defining the focus of the projection of a viewed prospect on a sensor that captures this projection. The size of the sensor is a physical constant and property of a photo camera, the so called focal length instead (that defines the distance between the sensor and the lens) usually varies with the changing of the optical zoom of the camera. The FOV of the captured image in this case is obviously the angular size of the part of the prospect projected on the sensor. Figure \ref{fig:fovSchema} shows this simple schema. We can easily write the relationship between the FOV of the photograph and the properties of the photo camera at the instant of shooting ($s$ for sensor width and $l$ for focal length): $$FOV = 2\arctan\frac{s}{2l}$$

Combining this definition with the previous relationship we can express the scaling factor $k$ that must be applied to the photograph in order to have the same object dimensions as: $$k = FOV\frac{w_{r}}{2\pi w_{p}} = \frac{w_{r}}{\pi w_{p}}\arctan\frac{s}{2l}$$

Scaling estimation is a purely mathematical procedure, and the quality of the results depends directly on the precision of the geotag and the accuracy of the rendered model (with exact GPS location and a render based on correct elevation data the scaling produces perfectly matchable images).

\subsection{Edge Detection}

\begin{figure}[h!]
	\includegraphics[width=1.0\columnwidth]{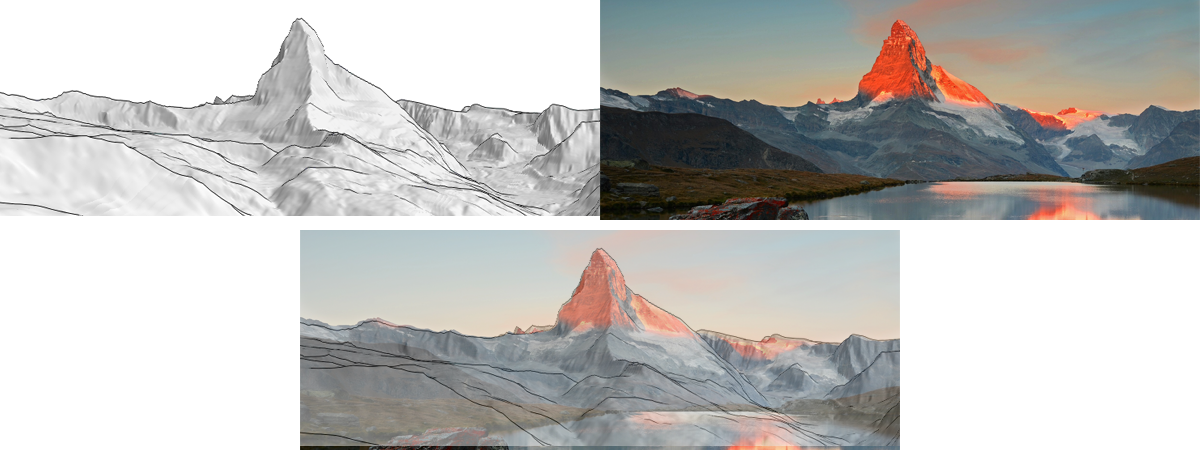}
	\caption{An example of a matching problem with the photograph fragment (top right), the panorama fragment (top left) and their overlapping (bottom).}
	\label{fig:mountainOverlapping}
\end{figure}

The key problem of the algorithm is to perform matching between the photograph containing the mountains and the automatically generated drawing representing the same mountain boundaries: in Figure \ref{fig:mountainOverlapping} we can see an example of a fragment of the photograph and a fragment of the rendered panorama. Both represent the same mountain seen from the same point of view, and in fact when we try to match them manually they overlap perfectly. In spite of this overlapping, the choice of the technique for their matching is not trivial. Even if the problem of matching (choosing the position of the photograph with respect to the panorama, maximizing the overlap between the mountains) will be discussed in the next sections, here we will briefly mention the possible techniques to evaluate the correctness of the overlap and the reasons for the necessity of the edge extraction procedure.

We can see the matching problem as a classic image content retrieval problem with the photograph as the input image and the collection of the fragments of the rendered panorama (each one corresponding to a possible alignment position) as the set of available images to search in. The most similar image in the set will be considered as the best matching position and identified as the direction of view of the camera during the shot.

So what is the similarity measure that can be used in order to perform this image content retrieval problem? Clearly global descriptors (color and texture descriptors) \cite{citeulike:10106398} are not the best choice: it is enough to look at the Figure \ref{fig:mountainOverlapping} to understand that the panorama is always generated in gray-scale tones, with textures defined only by the terrain elevation, while the mountains in the photograph can be colored in different ways, and the textures are defined by a lot of details such as snow on the mountains and other foreground objects such as grass, stones and reflections on the water. Local image descriptors instead (for example, local feature descriptors such as SIFT \cite{Lowe:2004:DIF:993451.996342} and SURF \cite{Bay:2008:SRF:1370312.1370556}) seem slightly suitable for the needs of mountain matching (as discussed also by Valgren et al. in \cite{Valgren:2010:SSS:1715935.1716080}, where the use of SIFT and SURF descriptors is highlighted for the matching of outdoor photographs in different season conditions), but even if they are good for matching the photographs containing the same objects in different color conditions, the matching between an object photograph and its schematic representation (in our case a rendered drawing) tends to fail. Even if very accurate, the model of a photograph will not generate local features with the same precision that another photograph of the same object would do: no local descriptors tested have been able to find the match between the two example images in Figure \ref{fig:mountainOverlapping}.

\begin{figure}[h!]
	\includegraphics[width=1.0\columnwidth]{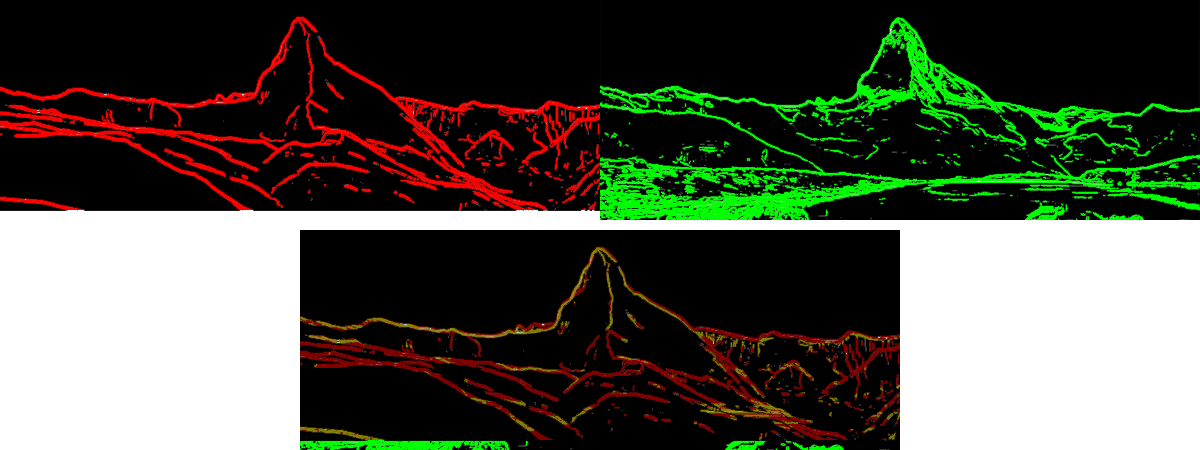}
	\caption{An example of a matching problem with the photograph edge fragment (top right), the panorama edge fragment (top left) and their overlapping (bottom).}
	\label{fig:mountainEdgesOverlapping}
\end{figure}

The perfect overlap between the images in the example figure however brings up to the idea, that instead of traditional image descriptors we should match the boundaries of the images (this assumption has also an intuitive motivation: the contours are the most significant and time invariant properties of the mountains), so the next step in photograph and panorama processing is edge extraction from both the images. The result of edge map extraction from the images in Figure \ref{fig:mountainOverlapping} is represented in Figure \ref{fig:mountainEdgesOverlapping}. The matching procedure on the edge maps can now be seen as a cross-correlation problem \cite{wiki:CrossCorrelation}.

In order to make the cross-correlation matching more sophisticated with a couple of techniques that will be presented later, the edge extraction component must accept as input an image (the photograph or the rendered panorama) and produce as output a 2D matrix of edge points, each one corresponding to the point on the input image with a strength (value between 0 and 1 representing the probability of the point to be an edge) and a direction (value between 0 and 2$\pi$ representing the direction of the edge in that point).

\subsection{Edge Filtering}
Once we have generated the edge maps of the photograph we must deal with the problem of the noise edge points. A noise edge point can be defined as an extracted edge point representing a feature that is not present on the rendered panorama, in this case the edge point will not be useful for the match and will only be able to harm it. In other words a noise edge point is an edge point that does not belong to a mountain boundary (the only features represented on our rendered panoramas). This doesn't mean that any edge point belonging to a mountain is not a noise point, i.e. the edge points that define the snow border of a mountain are noise points because they do not belong to the mountain boundary but define a border that due to its nature can easily change from photograph to photograph and furthermore will not be present in the rendered view.

However, edge points can be also present on the mountain surface; most of them usually belong to foreign objects. Mountains tend to be very edge-poor and detail-poor objects and often are placed in the background of a photograph with other edge-rich objects such as persons, animals, trees, or houses in the foreground. Let us think about a photograph of a person next to a tree with a mountain chain in the background: the mountains themselves will generate few edge points since they tend to be very homogeneously colored; foreground objects instead will generate a huge amount of edge points (which will be noise) because they are full of small details, each leaf of the tree and detail of the person's clothes will produce noise points.

\begin{figure}[h!]
	\centering
	\includegraphics[width=0.8\columnwidth]{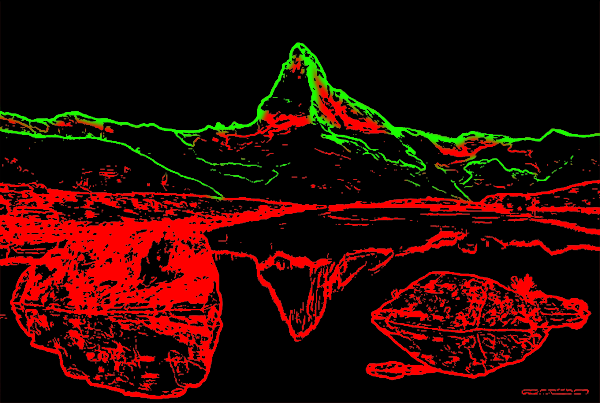}
	\caption{Noise (red) and non-noise (green) edge points on our example photograph.}
	\label{fig:filteringGoodBadBefore}
\end{figure}

Taking into account the example of the edge extraction made in the previous chapter, we manually tag the extracted points as noise and non-noise points following the edges present in the panorama: the result is represented in the Figure \ref{fig:filteringGoodBadBefore} and with a simple image analysis script we find that the the noise edge points are more than 90\% of all extracted points.
 This value grows up to 98\% when we deal with photographs with even more objects in foreground. Even if the matching algorithm we are going to propose includes penalizing the noise points in order to keep up with this problem the amount of the noise edges reaches such high levels that it cause almost any algorithm to fail simply from a statistical point of view (the intense density of noise points in some area tends to ``attract'' the matching position in that area), so an edge filtering technique is needed.
 
\begin{figure}[h!]
	\centering
	\includegraphics[width=0.8\columnwidth]{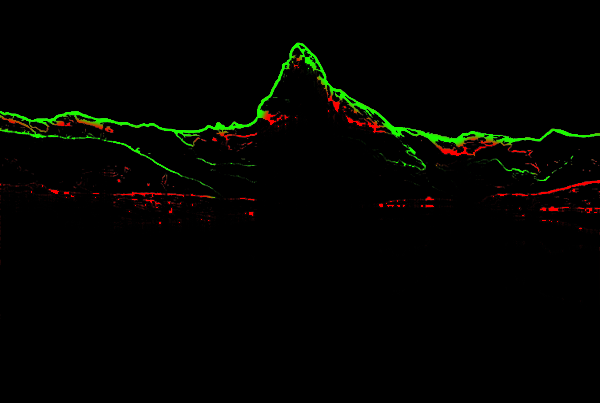}
	\caption{Noise (red) and non-noise (green) edge points on our example photograph after the filtering procedure has been applied.}
	\label{fig:filteringGoodBadAfter}
\end{figure}

One of the possible approaches is the detection of the skyline (the boundary between the sky and the objects present in the photograph) as implemented by Naval Jr et al. \cite{Jr97estimatingcamera} for the analogue problem. Skyline detection is a non-trivial task that has been widely studied with several proposed approaches available in particular sectors and scenarios, such as skyline detection of a perspective infrared image by synthesizing a catadioptric image proposed by Bazin et al. \cite{DBLP:conf/icra/BazinKDV09} or the extraction of the skyline from a moving omnidirectional camera as developed by Ramalingam et al. \cite{RBSB09}. The skyline however is not the only boundary of our interest, the boundaries of mountains contained ``inside'' other boundaries are also important and significant for the matching, so we decided to opt for softer a filtering approach.

We have opted for a simple but efficient technique based on the intuitive assumption that the mountains are usually placed above the other objects on the photographs (with some exceptions such as clouds or other atmospheric phenomena): prioritizing (increasing the strength of) the higher points on the photograph with respect to the lower points. An example of the result of the edge filtering procedure is displayed in Figure \ref{fig:filteringGoodBadAfter}. Most of the noise edges are filtered with almost all good edges intact (also the edges that would be cut with a skyline detection). The rate of the noise edge points dropped from 90\% to 40\% in this case.

\subsection{Edge matching (Vector Cross Correlation)}
\begin{figure}[h!]
	\centering
	\includegraphics[width=1.0\columnwidth]{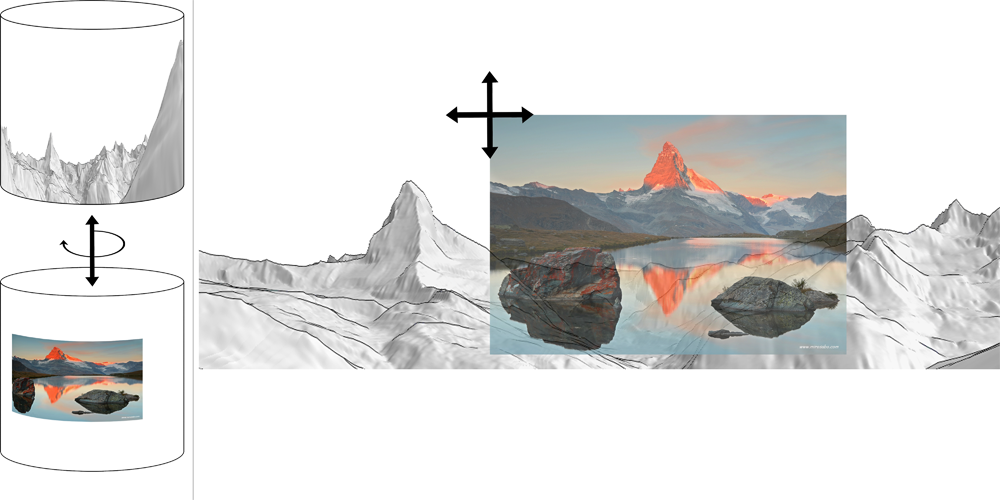}
	\caption{Representation of cylindrical image matching and the equivalent rectangular image matching.}
	\label{fig:cylinderMatching}
\end{figure}

Firstly we define $C_{r}$ as the cylindrical image generated from the rendered panorama with the same height, and the base perimeter equal to the panorama's width, and $C_{p}$ as the input photograph projected on an empty cylindrical image of the same dimensions of $C_{r}$. Imagining two cylinders to be concentric the matching problem is defined as the two dimensional space search (the vertical position and the rotation angle of $C_{r}$ with respect to $C_{p}$), which leads to the best overlap between the mountains present on the cylindrical images. As the cylindrical images are only a projection of a rendered image on a cylinder, the problem can be equivalently seen as a search for the best overlap of two rectangular images defined by two integer numbers representing the coordinate offset of the photograph with respect to the panorama (obviously carrying out the horizontal overflow part of the photograph to the opposite side of the render, simulating a cylinder property) as shown in Figure \ref{fig:cylinderMatching}. 

As already introduced in previous sections, the matching will be performed with the edge maps of the images so we define the result of the edge detection algorithm as a 2D real-valued vector where each value is defined by $\rho$ (the strength/absolute value of the edge in the corresponding point of the input image) and $\theta$ (the direction/argument of that edge). Let $p(\omega)$ and $r(\omega)$ be the 2D real-valued vectors generated by edge detection of the photograph and the panorama render respectively. Then as proposed by Baboud et al. \cite{Baboud2011Alignment} the best matching is  defined as the position maximizing the likelihood between two images. This likelihood is defined as $$L(p,r) = \int_{S^{2}} M(p(\omega),r(\omega))d\omega$$
where $M$ is the angular similarity operator:
$$M(v_1,v_2) = \rho ^{2}_{v_1} \rho ^{2}_{v_2} \cos 2(\theta_{v_1} - \theta_{v_2})$$

\begin{figure}[h!]
	\centering
	\includegraphics[width=1.0\columnwidth]{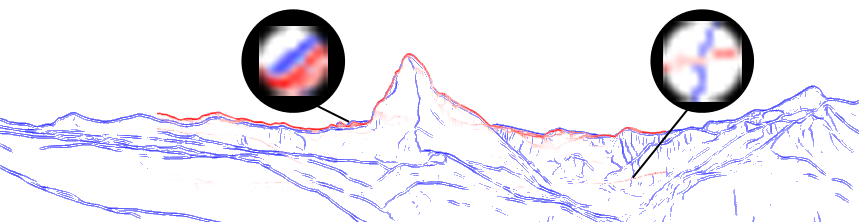}
	\caption{Example of an overlapping position with positive score with almost parallel edge intersection (left circle) and penalizing almost perpendicular edge intersection (right circle).}
	\label{fig:penalizingEdges}
\end{figure}

This technique of considering the edge maps as vector matrices and applying the cross-correlating resolution is called Vector Cross Correlation (VCC). The cosine factor is introduced in order to handle edge noise by penalizing differently oriented edges: the score contribution is maximum when the orientation is equal, null when they form a $\frac{\pi}{4}$ angle, and minimum negative when the edges are perpendicular (Figure \ref{fig:penalizingEdges}). This penalty avoids that  random noise edges contribute in a positive way to a wrong match position (a step in this direction was already made during the edge filtering).

One of the main advantages of this likelihood score is the possibility of applying the Fourier transform in order to reduce drastically the computation effort of best matching position: let us define $\hat{p}$ and $\hat{r}$ as 2D Fourier transforms of respectively $p(\omega)$ and $r(\omega)$. The VCC computation equation becomes
\begin{equation} \label{eq:FFTVCC}
L(p,r) = Re\{\hat{p}^{2} \bar{\hat{r}}^{2}\}
\end{equation}
This cosine similarity VCC can be seen as the whole algorithm to perform the matching or only as its first step: if we build the 2D real-valued distribution of the likelihood values estimated in each possible position we can consider not only the best match but extract top-N peaks of the score distribution as the result candidates and then evaluate them with a more sophisticated and heavier technique in order to identify the correct match. One of these refining algorithms is the robust silhouette map matching metric technique designed by Baboud et al. \cite{Baboud2011Alignment} which considers the edge maps as sets of singular connected edges. The likelihood of an overlap is the sum of the similarity between each edge $e_p$ of the photograph and the edge $e_r$ of the rendered panorama, where $e_r$ is enriched with a certain $\epsilon_e$ neighborhood, $l$ represents the distance for which $e_p$ stays inside the $\epsilon_e$ neighborhood of the $e_r$ and the similarity is defined as
\begin{equation} \label{eq:robustMatching}
M(e_p, e_r) = 
\left\{\begin{matrix}
0
\\ 
l^a
\\ 
-c
\end{matrix}\right.
\begin{array}{ll}
\text{if }l = 0
\\ 
\text{if }(l > l_{fit}) \land (e_p \text{ enters and exits on the same side})
\\ 
\text{if }(l < l_{fit}) \land (e_p \text{ enters and exits on different sides})
\end{array}
\end{equation}
where $a$, $c$ and $l_{fit}$ are predefined constants. The nonlinearity implied by the exponent $a$ makes longer edge overlaps receive more weight than the set of small overlaps, the constant $c$ instead introduces also in this metric the penalizing factor for the intersections of the edges considered wrong as in the VCC metric.


\subsection{Mountain Identification and Tagging}
\begin{figure}[h!]
	\centering
	\includegraphics[width=1.0\columnwidth]{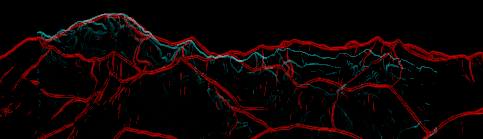}
	\caption{Example of a fragment of a matching result between the photograph (blue) and the panorama (red).}
	\label{fig:resultWrongPeaks}
\end{figure}

\begin{figure}[h!]
	\centering
	\includegraphics[width=1.0\columnwidth]{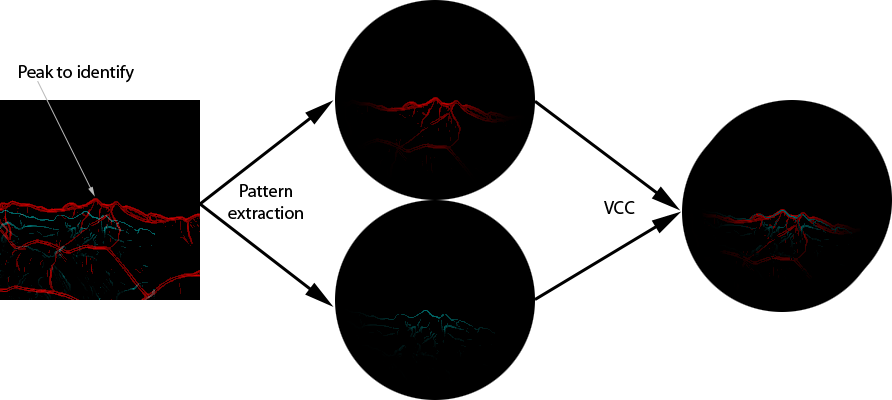}
	\caption{Peak identification procedure on the previous matching example.}
	\label{fig:peakIdentification}
\end{figure}

Once the edges are matched and the direction of view of the camera during the shot is estimated, supposing to have the list of mountain peaks with their names and coordinates on the rendered panorama we can estimate the position of these peaks also on the photograph, even if it is not as trivial as can be initially thought. Intuition suggests that once the photograph is matched with the panorama the coordinates of the mountain peaks are the same on both (obviously with a fixed offset equal to the matching position in case of the photograph) but it is wrong. Looking at a fragment of the result of edge matching (the best estimated position by VCC) of a photograph in Figure \ref{fig:resultWrongPeaks}: the overlapping between the photograph and panorama edges is almost perfect on the left part but tends to be always worse moving to the right, which means that the two edge maps are slightly different (probably due to the error in position and altitude estimation) and objectively the matching proposed by the VCC seems to be the best that can be obtained. This situation occurs very often, and in general we can say that the resulting edge matching, even when successful, presents small errors in singular mountain peaks due to the imperfections of the generated panorama model. We propose a method for precise mountain peak identification: for each mountain peak present in the panorama we extract an edge map pattern both from the photograph and the panorama centered in the coordinate of the peak on the panorama and weighted with a certain kernel function $f(d)$ for a fixed pattern radius $r$ and $d$ representing a point distance from the peak coordinate with the following properties:
$$
\begin{array}{ll}
f(0) = 1
\\ 
f(x) = 0 \text{ }
\\ 
f(x_1) \geq f(x_2) \text{ }
\end{array}
\begin{array}{ll}

\\ 
\forall x \geq r
\\ 
\forall x_1, x_2 \, \mid \, x_1 \leq x_2
\end{array}
$$
Once the two patterns are extracted the VCC procedure is once again applied in order to match them, having only the edges of the mountain peak we are processing and a few surrounding edges, the matching position is no longer influenced by the other peaks and is refined to the exact position, an example of this procedure performed on the matching result introduced in Figure \ref{fig:resultWrongPeaks} in order to identify one of the most right-placed peaks is displayed in Figure \ref{fig:peakIdentification}.

\chapter{Implementation Details}
\label{Implementation Details}
\thispagestyle{empty}

\vspace{0.5cm}

\noindent
In this chapter the implementation of the matching algorithm proposed in the previous chapter is presented, from the general description of the system architecture and programming languages involved to the value of each parameter and constant introduced in the algorithm presentation.

\section{Overview of the implementation architecture}
The whole system can be split into two macro areas:
\begin{itemize}
\item
\emph{Photograph analysis, panorama and mountain peak list generation}: implemented in PHP with web interface due to the simplicity of interfacing with the external panorama generation service and interface versatility. Given a geo-tagged photograph (or a photograph with explicitly specified the geographic point of the shot) the virtual panorama view is generated centered at the same origin point and the properties both of the photograph and the photo camera are retrieved in order to let the next step calculate the field of view and scale factor.
\item
\emph{Matching and mountain identification}: implemented in MATLAB for the high suitability with image processing techniques. Given a photograph with all necessary information, the panorama and the mountain list with the relative coordinates on the panorama, edge extraction and filtering is performed, the camera direction is estimated, and finally individual mountains are identified and tagged.
\end{itemize}

The choice of the operating parameters used in the final implementation was defined in the testing and validation phase, using a data set and a developed evaluation metric. Although the parameter values will be specified as soon as they are introduced, the detailed description and the reasons of discarding the other values will be presented in the next chapter.

\section{Detailed description of the implementation}

\subsection{Render Generation}
\begin{figure}[h!]
	\centering
	\includegraphics[width=1.0\columnwidth]{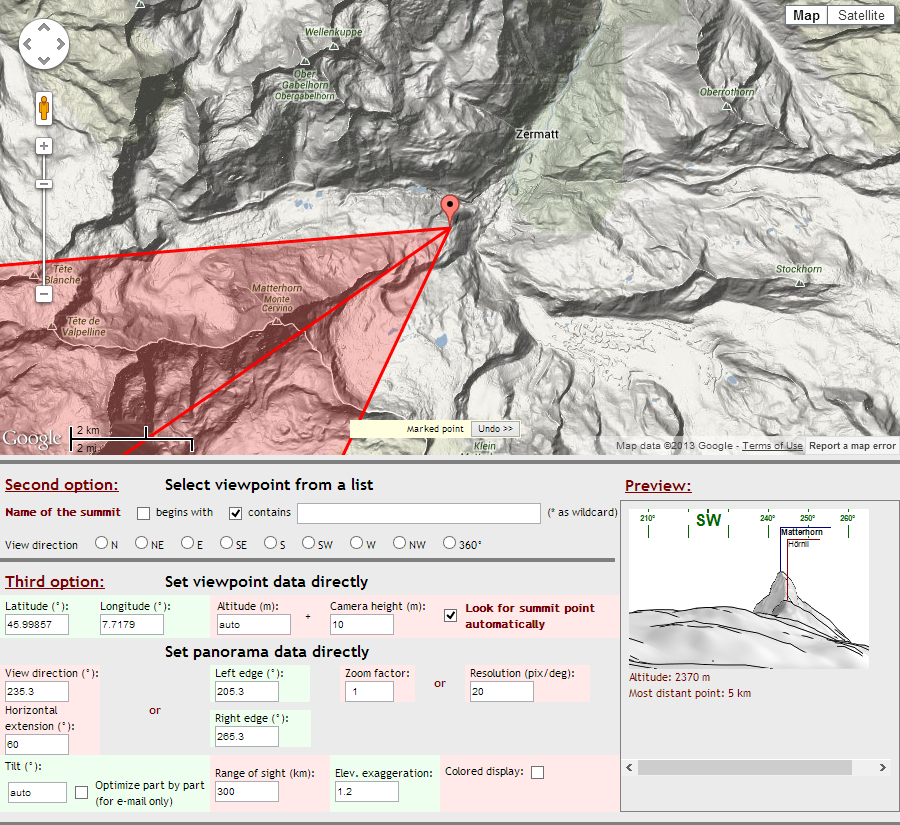}
	\caption{A screenshot of the mountain panorama generation web service interface.}
	\label{fig:udeuschleScreenshot}
\end{figure}

As already mentioned in the implementation an external panorama service has been used: the mountain view generator of Dr. Ulrich Deuschle \footnote{\url{http://www.udeuschle.de/panoramen.html}} (Figure \ref{fig:udeuschleScreenshot}). The service accepts as input several parameters, most important of which are the geographic position of the observer, his altitude and altitude offset, the view direction with the field of view, zoom factor, elevation exaggeration and the range of sight. Latitude and longitude are equal obviously to the geo-tag of the input photograph, and the horizontal extension (field of view) to the round angle, $2\pi$.

The choice of the altitude (in fact of the altitude offset) as already anticipated in the previous chapter is a problematic point: a big offset can lead to vertical distortion of the panorama with respect to reality, a small offset on the other hand can lead more easily to partial occlusion of the panorama. The choice was made for $auto + 10$~m altitude setting, chosen by trial and error. Regarding the altitude choice an important option of the service named ``Look for summit point automatically'' must be highlighted: by turning on this option the panorama is generated from the highest point of the terrain within 200~m. The advantage/disadvantage of using this option is the same as increasing the altitude offset: avoiding the occlusion of the panorama against distorting the generated view. Even if this option is very interesting we set it to off as the distortion of the panorama due to the observer's position changing and not only his altitude was too detrimental for the algorithm.

For the most realistic panorama the zoom factor and elevation exaggeration factors are both set to $1$. All the other parameters are left to their default values.

This service covers the the Alps, Pyrenees and Himalayas mountain range systems with two elevation datasets:
\begin{itemize}
\item
\emph{Alps} by Jonathan de Ferranti, with the spatial resolution of $1$~arcseconds
\item
\emph{SRTM} by CGIAR, with the spatial resolution of $3''$
\end{itemize}

The output is implemented with dynamic loading of the panorama, so is generated as a set of images to be aligned horizontally to form the complete output image and the one image containing the mountain peak names to be overlapped with the result image. Despite the unavailability of the service API, after the study of JavaScript and AJAX scripts of the web interface, a PHP function that simulates the functioning of the interface and collects all the parts of the result image was implemented.

The generation of the full panorama at the resolution of $20$~pixel/degree and the peaks names takes on average 1 minute.

\subsection{Field of View Estimation and Scaling}
To estimate the field of view of the input photograph, as described in the previous chapter, the only information that must be known is the focal length of the photograph and the sensor width of the photo camera. The focal length is a parameter specified directly in the EXIF format of the image (tag = FocalLength, 37386 (920A.H)) \cite{Technical2002Exchangeable}. The sensor width instead is not specified directly in the EXIF since it is a property of a photo camera, so the manufacturer and the model of the camera are extracted from the EXIF and then the sensor size is retrieved (manufacturer tag = Make, 271 (10F.H); model tag = Model, 272 (110.H)) \cite{Technical2002Exchangeable}. 

Clearly, the necessity of a database of camera sensor sizes emerges. This information is usually scattered on web sites of manufactures and technical references of the cameras, so it is not easy to collect them in one place. We used a database kindly provided by ``Digital Camera Database'' \footnote{\url{http://www.digicamdb.com}} web site, containing information about the sensor size of more than 3000 digital cameras. The main problem in using it remains in the fact that the manufacturer and model names of the same cameras are not always equal between that written in the EXIF and that provided by the sensor database, a couple of these mismatching examples are provided in Table \ref{tab:exifMismatching}.

\begin{table}[!h]
  \centering
  \begin{tabularx}{\textwidth}{|X|X|X|X|}
    \hline
    \multicolumn{2}{|c|}{\tabhead{EXIF}} &
    \multicolumn{2}{c|}{\tabhead{Database}} \\
    \hline
    \tabhead{Manufacturer} &
    \tabhead{Model} &
    \tabhead{Manufacturer} &
    \tabhead{Model} \\
    \hline
    Canon &
    Canon PowerShot SX100 IS &
    Canon &
    PowerShot SX110 IS \\
    \hline
    SONY &
    DSC-W530 &
    Sony &
    Cybershot DSC W530 \\
    \hline
    NIKON &
    E5600 &
    Nikon &
    Coolpix 5600 \\
    \hline
    OLYMPUS IMAGING CORP. &
    SP560UZ &
    Olympus &
    SP 560 UZ \\
    \hline
  \end{tabularx}
  \caption{Examples of differences in the names of manufacturers and models between the EXIF specifications and the digital camera database.}
  \label{tab:exifMismatching}
\end{table}

To find the correct photo camera in the database from the EXIF specifications name a text similarity score between the names is used and the most similar name is chosen. The text similarity is calculated by the \emph{similar\_text} PHP function proposed by Oliver \cite{DBLP:books/daglib/0077674}, after several steps of preprocessing:

\begin{enumerate}
\item
Both the manufacturer and model names of both the EXIF and the database items are transformed to lower case.
\item
If the manufacturer name contains ``nikon'' the name is set to ``nikon''.
\item
If the manufacturer name contains ``olympus'' the name is set to ``olympus''.
\item
If the model name contains the manufacturer name it is cut off from the model name.
\item
The text similarity score is performed between the concatenation of the manufacturer and the model of both EXIF and database items.
\end{enumerate}

Once the focal length and the sensor size are retrieved, they are annotated within the image, the matching algorithm will later use them to calculate the field of view and scale factor.

\subsection{Edge Detection}
For the edge extraction the \emph{compass} \cite{Ruzon:2001:EJC:505471.505477} edge detector has been used. It returns exactly the output the matching algorithm needs (for each point the edge strength and direction) and has been chosen due to its ability of exploiting the whole color information contained in the image and not only the grayscale components as classical edge detectors do.
The \emph{compass} detector deals well also with a significant problem of edge detectors: when the image presents a subjective boundary between two regions with pixels close in color (due to overlapping objects), most edge detectors compute a weighted average of the pixels on each side, and since the two values representing the average color of two regions are close, no edges are found \cite{Ruzon:2001:EJC:505471.505477}. The chosen detector deals well with this problem, that is very likely in the context of mountain photographs and the boundaries between the snowy mountain and blue sky.

The edge detection procedure is applied both to the input photograph and the generated panorama, with the following parameters chosen by trial-and-error: standard deviation of Gaussian used to weight pixels $\sigma = 1$, threshold of the minimum edge strength to be considered $\tau = 0.3$.

\subsection{Edge Filtering}
The edge filtering approach used in the implementation treats the columns of the image separately: each column is split into segments separated by the zero strength edge points, and each segment is then split into segments of length $n$ points. The points of each $i$-th segment (starting from the top) are then multiplied by the factor of $b^{i-1}$.

\begin{figure}[h!]
	\centering
	\includegraphics[width=1.0\columnwidth]{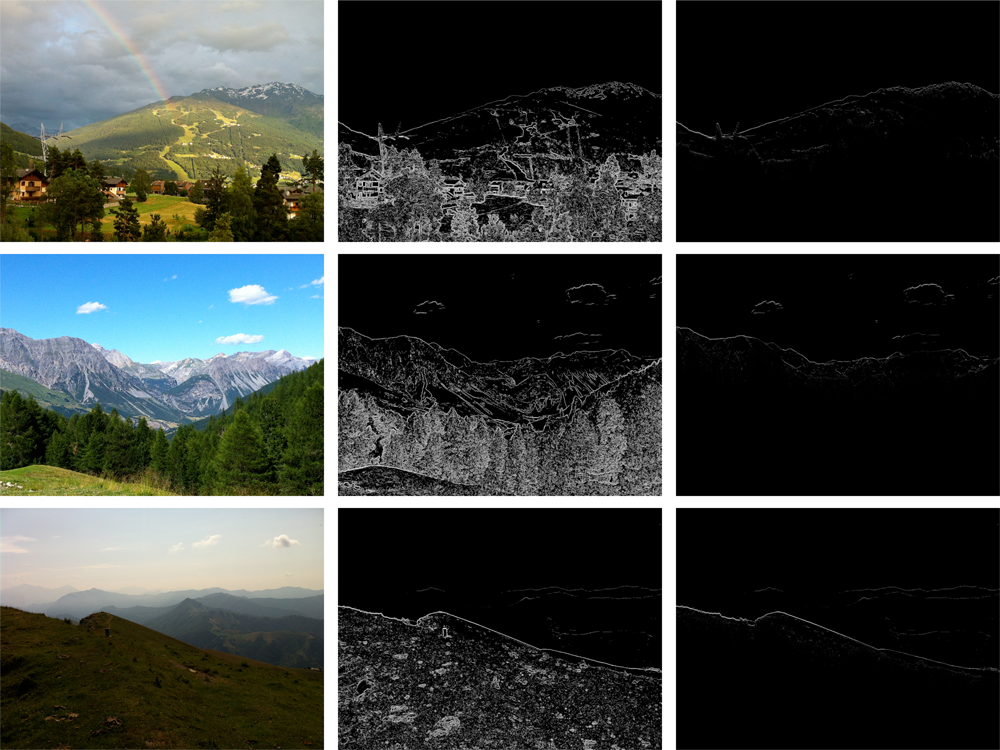}
	\caption{Examples of edge filtering (one for each row): the original photograph (left), the extracted edges (center) and the filtered edges (right).}
	\label{fig:filteringExamples}
\end{figure}

The implementation uses $k = 2$ and $b = 0.7$. It allows a good filtering of noise objects at the bottom of the photograph, and meanwhile deals well with mountain edges placed below the clouds. Several examples of filtering are presented in Figure \ref{fig:filteringExamples}:
\begin{itemize}
\item
\emph{First row}: a photograph with clouds and a rainbow completely removed from the edges map (thanks to the correct parameters of the edge extraction algorithm).
\item
\emph{Second row}: a photograph with strongly contrasting clouds, the edges of the mountains below the clouds are still present even if reduced in strength.
\item
\emph{Third row}: a photograph with different mountains, one in front of another with a reduced visibility, even if reduced in strength all the mountains are present on the filtered edge map and the noise edges of the terrain in the foreground are filtered.
\end{itemize}

\subsection{Edge matching (Vector Cross Correlation)}
With the edge maps of the photograph and the panorama the matching of the edges is performed by fast Fourier transformation, with the Formula \ref{eq:FFTVCC}. The MATLAB implementation of the VCC is the following:

\begin{lstlisting}[frame=single]
function VCC = ComputeVCC(SP, DP, SR, DR)

% VCC = ComputeVCC(SP, DP, SR, DR)
%
% This function takes a real valued matrices:
% 
% SP - strength matrix of the edge points of the input photograph
% DP - direction matrix of the edge points of the input photograph
% SR - strength matrix of the edge points of the rendered panorama
% DR - direction matrix of the edge points of the rendered panorama
% 
% and returns the resulting VCC matrix with the overlapping score
% for each possible position

% An empty fragment is added to the top and bottom of rendered panorama
% to reduce the circular cross-correlation to a cylindrical one
dimP = size(SP);
dimR = size(SR);
SR = [ zeros(dimP(1), dimR(2)) ; SR ; zeros(dimP(1), dimR(2)) ];
DR = [ zeros(dimP(1), dimR(2)) ; DR ; zeros(dimP(1), dimR(2)) ];

COMP = complex( SP .* cos(DP) , SP .* sin(DP) );
COMR = complex( SR .* cos(DR) , SR .* sin(DR) );

VCC = rot90(real(ifft2(conj(fft2(COMR.^2)) .* fft2(COMP.^2, size(COMR, 1), size(COMR, 2)))),2);

VCC = VCC( dimP(1) + 1 : 2 * dimP(1) );
\end{lstlisting}

\begin{figure}[h!]
	\centering
	\includegraphics[width=1.0\columnwidth]{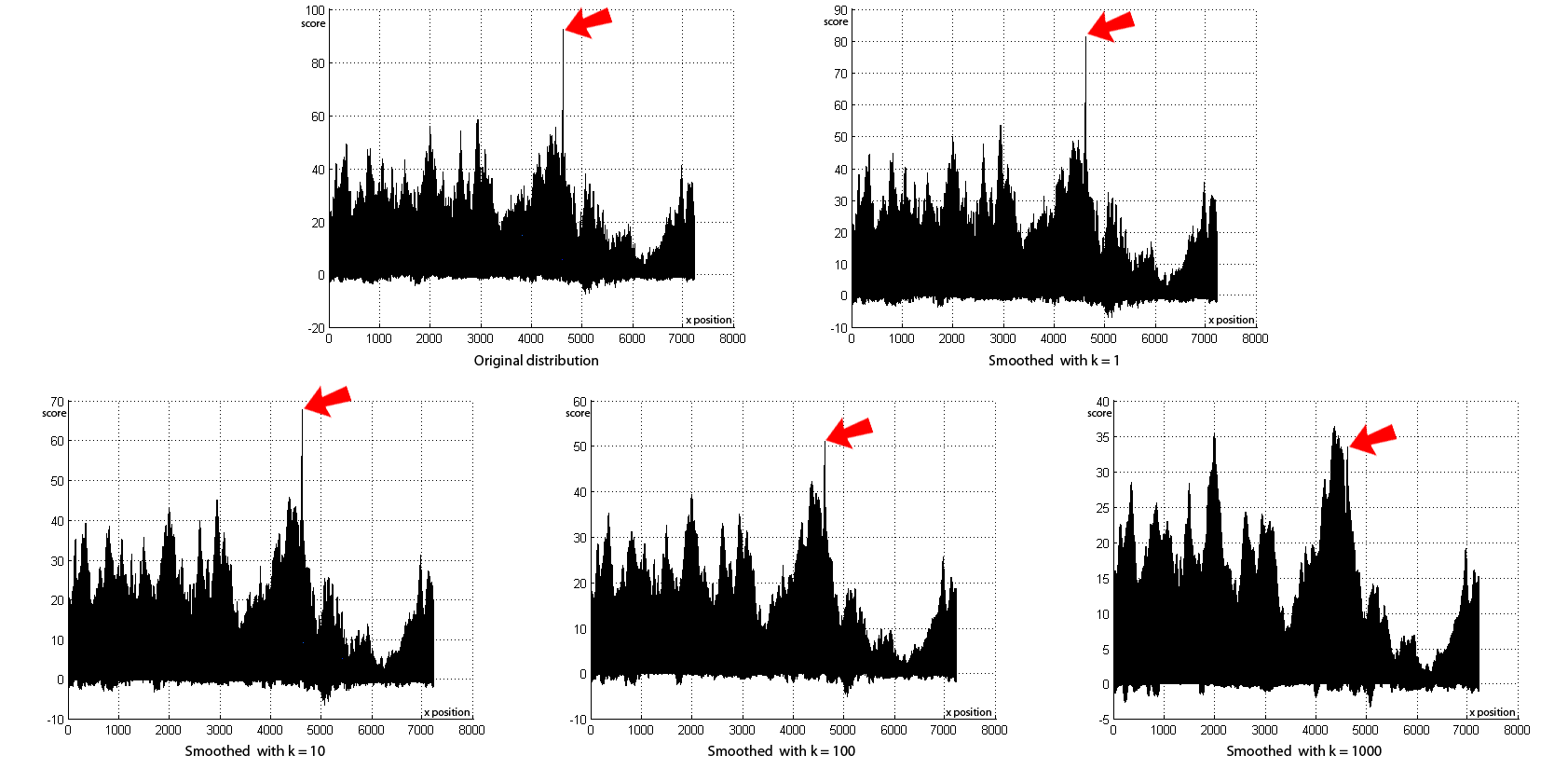}
	\caption{Smoothing of the VCC result distribution with smoothing factor $k$. The red arrow identifies the correct matching.}
	\label{fig:smoothing}
\end{figure}

The resulting matrix of this function is the VCC score evaluation for each possible overlapping position, and the maximum value of the matrix identifies the best overlap. Instead of just peaking up the maximum value, several additional techniques have been implemented and tested:

\emph{Result smoothing}: the idea is that the result score peak will have high values also in the neighborhood of its position, and the noise score peaks instead do not: in other words the peak corresponding to the right position will have be broader with respect to noise score peaks. So smoothing the result score distribution penalizes the noise peaks: it will reduce more the height of the noise peaks with respect to the correct peak. In practice, after several tests, the incorrectness of this assumption emerges. Due to the nature of the VCC technique of penalizing the non parallel intersection of the edges, even if the score in the correct position is high, it is sufficient to move only few points out to reduce drastically the score, so the smoothing procedure was not efficient. Figure \ref{fig:smoothing} shows an example of the score matrix (projected onto the $X$--$Z$ plane for readability) of the photograph of the Matterhorn used in the previous examples: the correct position represents already the maximum of the distribution, and not only it is intuitively visible that the correct position score peak is not smoother than the others, but it is also shown that more smoothing always leads to a smaller difference between correct and incorrect matches. 

\emph{Robust matching}: as proposed in the algorithm chapter, the top-$N$ peaks of the score matrix are extracted, each one is evaluated to find the best matching position. The evaluation is performed with Formula \ref{eq:robustMatching}, that has been implemented in a simplified version in the following way:
\begin{enumerate}
\item
The information about the direction of the edge points is removed, all the edge points with strength bigger than $0$ are considered to have the strength equal to $1$.
\item
On the panorama edge map each point that is located less or equal than $r$ points from any edge point in the original edge map is set to $1$. In this way the $r$-neighborhood of the panorama edges is generated.
\item
A simple intersection between the new panorama edge map and the photograph edge map is performed in the overlapping position that must be evaluated.
\item
The resulting intersection (composed only by $0$ and $1$ strength points) is divided into clusters where any point of the cluster is located less than $d$ from at least one point of the same cluster.
\item
The Formula is applied where the clusters are singular edges, and the edge length is the number of the points in the cluster.
\end{enumerate}

After performing tests on the dataset also this approach has been rejected as increasing drastically the computational effort of the matching without introducing significant benefits to the results.

\emph{Different scale factors}: since the reduced times of VCC computation with the fast Fourier transform ($\sim 1$~second for the VCC score matrix generation in our implementation) an approach to reduce the impact of the wrong photograph shot position or the incorrect field of view estimation (due for example to a wrong sensor peaking) is to perform the matching at several scale levels and not only the estimated scale. In the current implementation several scaling intervals with respect to the estimated level were evaluated, and the scale factor with the best VCC maximum value was picked. Obviously a larger scale factor will lead to a larger photograph edge map and so bigger VCC score, so when comparing different maxim values an inverse quadratic scale factor must be applied to each score (inversely proportional to the area added to the image due to the scale changing).

\subsection{Mountain Identification and Tagging}

\begin{figure}[h!]
	\centering
	\includegraphics[width=1.0\columnwidth]{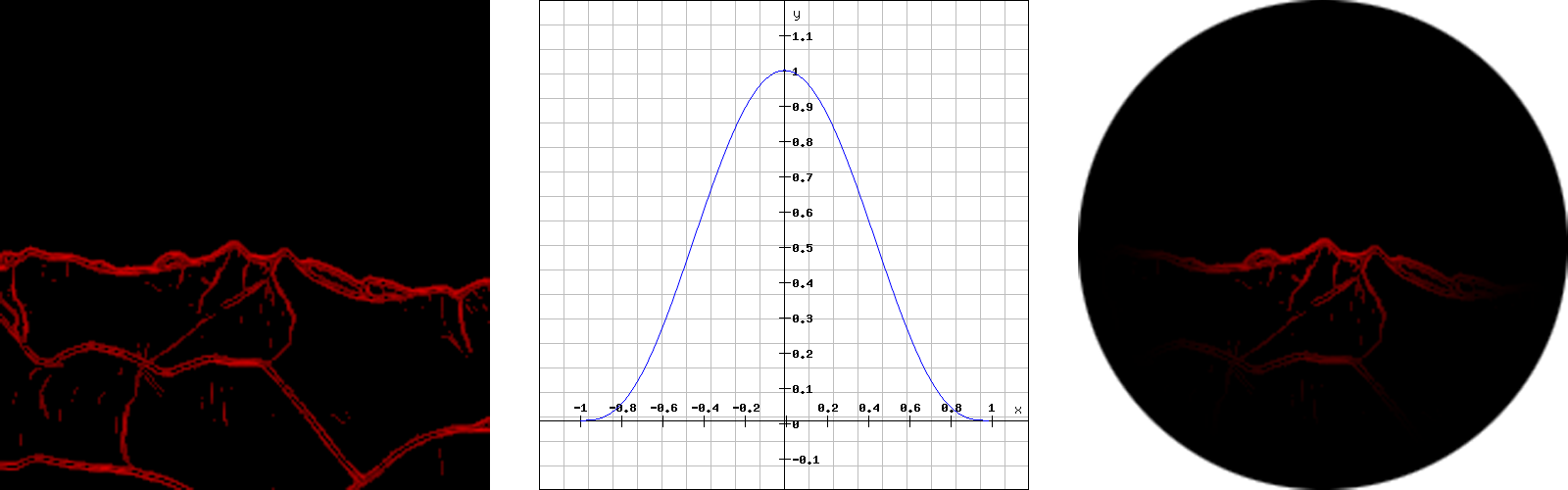}
	\caption{Peak extraction: edge map before pattern extraction (left), implemented kernel function with $r = 1$ (center), edge map after pattern extraction (right).}
	\label{fig:filteringFunction}
\end{figure}

The kernel function chosen for the peak extraction for the identification is the triweight function, defined as:

$$
f(d) = 
\left\{\begin{matrix}
\left(1-\left(\frac{d}{r}\right)^{2}\right)^{3} \\ 
0
\end{matrix}\right.
\begin{matrix}
\text{ for }d \leq r \\
\text{ for }d > r
\end{matrix}
$$

With $r = 200$. The kernel function plot and its effect on the peak extraction are shown in Figure \ref{fig:filteringFunction}.

\chapter{Experimental Study}
\label{Experimental Study}
\thispagestyle{empty}

\vspace{0.5cm}

In this chapter the photograph direction estimation quality is investigated. The precision of the estimation is computed varying the operating parameters of the implementation and the type of the photographs contained in the data set.

\section{Data sets}
\begin{figure}[h!]
	\centering
	\includegraphics[width=1.00\columnwidth]{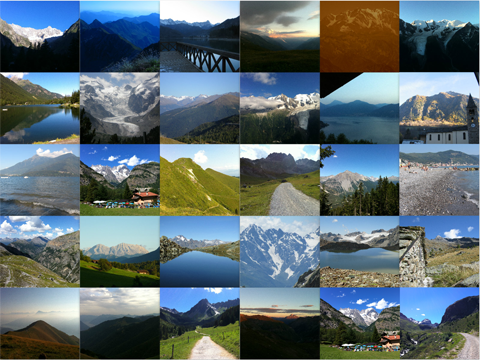}
	\caption{Several photographs from the collected data set.}
	\label{fig:dataset}
\end{figure}

The analysis is conducted on a set of 95 photographs of the Italian Alps, collected from several photographers with geo-tag set directly by a GPS component or manually by the photographer, so we consider them very precisely geo-tagged. The photographs vary in several aspects: several examples are shown in Figure \ref{fig:dataset}, and the categories of interest are composed as described in Table \ref{tab:datasetCategoriesComposition}.
\begin{table}[!h]
  \centering
  \begin{tabularx}{\textwidth}{|X|c|c|}
    \hline
    \tabhead{Category} &
    \tabhead{Option} &
    \tabhead{Data set portion} \\
    \hline
    \multirow{2}{*}{Source} &
    Photo camera &
    38 \% \\
    &
    Cellular phone &
    62 \% \\
    \hline
    \multirow{2}{*}{Cloud presence} &
    None &
    41 \% \\
    &
    Minimal &
    29 \% \\
    &
    Massive &
    23 \% \\    
    &
    Overcast &
    7 \% \\    
    \hline
    \multirow{2}{*}{Skyline composition} &
    Mountains and terrain only  &
    87 \% \\
    &
    Foreign objects &
    13 \% \\
    \hline    
  \end{tabularx}
  \caption{Data set categories of interest composition.}
  \label{tab:datasetCategoriesComposition}
\end{table}

\section{Operating parameters}
The operating parameters used in the algorithm described in the previous sections, and to be evaluated, are listed in Table \ref{tab:datasetOperatingParameters}. The value in bold defines the default value that gives the best evaluation results and is used in the final implementation proposal.

\begin{table}[!h]
  \centering
  \begin{tabularx}{\textwidth}{|X|c|c|}
    \hline
    \tabhead{Full name} &
    \tabhead{Parameter} &
    \tabhead{Tested values} \\
    \hline
    Photograph edge strength threshold &
    $\rho_p$ &
    0.1,0.2,\textbf{0.3},0.4,0.5,0.6,0.7 \\
    \hline
    Panorama edge strength threshold &
    $\rho_r$ &
    0.1,\textbf{0.2},0.3,0.4,0.5 \\
    \hline    
    Photograph edge filtering base &
    $b_p$ &
    0.5,0.6,\textbf{0.7},0.8,0.9,1.0 \\
    \hline    
    Panorama edge filtering base &
    $b_r$ &
    0.5,0.6,0.7,0.8,0.9,\textbf{1.0} \\
    \hline    
    Photograph edge filtering max segment length &
    $l_p$ &
    1,\textbf{2},3,4,5 \\
    \hline    
    Panorama edge filtering max segment length &
    $l_r$ &
    1,2,3,4,5 \\    
    \hline    
    Photograph scaling interval &
    $\pm k \%$ &
    \textbf{0},1,2,5,10 \\    
    \hline           
  \end{tabularx}
  \caption{Operating parameters (defaults in bold).}
  \label{tab:datasetOperatingParameters}
\end{table}

\section{Evaluation metric}
Each photograph in the data set and the corresponding generated panorama are manually tagged by specifying on both one or more pairs of points corresponding to the same mountain peaks. Once the direction of view is estimated and the best overlap of the edges is found, the error of the direction estimation is defined as the average distance between the position of the peak on the panorama and the position of the peak on the photograph projected on the panorama with the current overlap position. The distance is then expressed as the angle, considering the panorama width as $2\pi$ angle. The error therefore lies between 0 and just over $\pi$ angle (the worst horizontal error is $\pi$ in the case of opposite directions, but since the vertical error is also counted the total error can theoretically slightly exceed this value).

Given a photograph and a panorama rendered with the resolution of $q$ pixel/degree, both tagged with $N$ peaks, let $x_{pi}$/$y_{pi}$ and $x_{ri}$/$y_{ri}$ be respectively the coordinates (expressed in pixels) of the $i$-th peak on the photograph and the panorama image, the estimation error is defined as:

$$
e = \frac{\sqrt{(\sum_{i = 1}^{N} (x_{pi} - x_{ri}))^2 + (\sum_{i = 1}^{N} (y_{pi} - y_{ri}))^2 }}{Nq}
$$

\begin{figure}[h!]
	\centering
	\includegraphics[width=1.00\columnwidth]{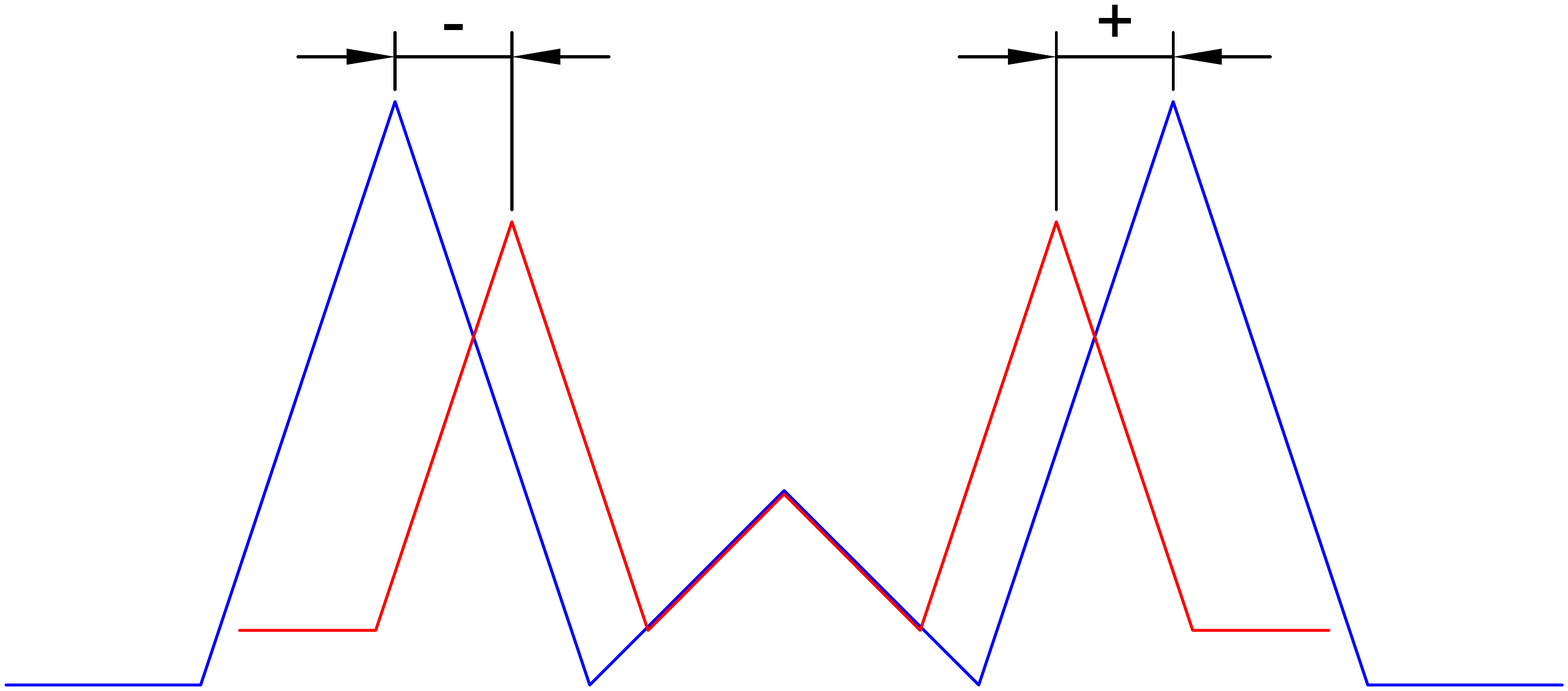}
	\caption{Schematic example of edge matching validation with photograph edges (red) and panorama edges (blue), all the three peaks are validation points.}
	\label{fig:errorOffsetValidation}
\end{figure}

An important aspect of the average computing must be highlighted: instead of the standard distance error averaging by taking into account the singular distances, this error metric calculates the average of the both image coordinates axis components, and then the distance with these components is computed. In practice, a positive error offset compensates a negative one. This is made to compensate the imperfections between the scale of the mountains on the photograph and panorama: if the overlap presents both positive and negative offsets between peak pairs it means that the photograph and the panorama are differently scaled, so the best overlap position is the one that brings the sum of the offset components to zero. An example of this reasoning is shown as a schematic case of edge matching in Figure \ref{fig:errorOffsetValidation} the validation points include all three peaks of the photograph and panorama view, but the different scaling prevents perfect overlapping and the offsets are both positive and negative. Thus the best possible overlapping position is the one shown, that makes the offsets opposite and making the sum of the offsets converge to zero.

Therefore even when the validation points cannot all be matched perfectly, the technique of not taking the absolute value of the offsets makes it anyway possible to find the best match and obtain an offset of zero (perfect score).

For readability reasons all the validation errors from this point will be specified in arc degrees ($\degree$).

Once the matching error is calculated for every photograph present in the data set, the general score of the run of the algorithm is defined as the percentage of the portion of the data set containing all the photographs with matching error below a certain threshold. This threshold in other words defines the limit of the matching error for a photograph to be considered matched correctly, and in the evaluation of this work is set to $4\degree$.

\section{Evaluation results}
\begin{figure}[h!]
	\centering
	\includegraphics[width=1.00\columnwidth]{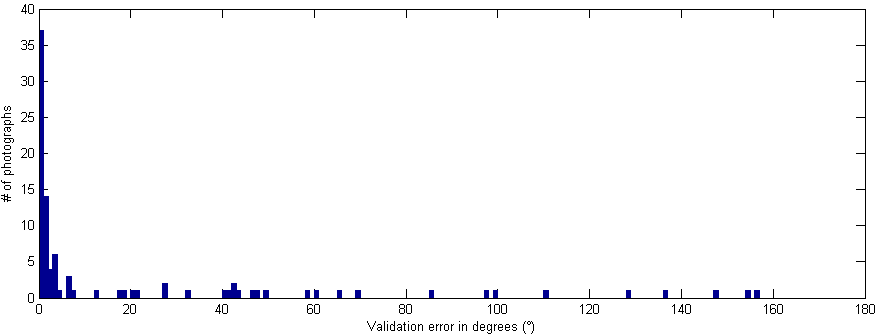}
	\caption{Histogram of the number of the photographs with a certain validation error (split into intervals of width $1$~\degree.)}
	\label{fig:mainResultHist}
\end{figure}

With the all operating parameters set to their default values the algorithm has correctly estimated the orientation of \textbf{64.2\%} of the total photographs. In particular the distribution of the matching error through the photographs of the data is presented in Figure \ref{fig:mainResultHist}: it is clear that the choice of the error threshold has a limited impact on the correct matching rate since almost all the correctly matched photographs have error between $0\degree$ and $2\degree$.

The results obtained with the described data set decomposed into categories as described in Table \ref{tab:datasetCategoriesComposition} are summarized in Table \ref{tab:datasetCategoriesResults}.

\begin{table}[!h]
  \centering
  \begin{tabular}{|c|c|c|}
    \hline
    \tabhead{Category} &
    \tabhead{Option} &
    \tabhead{Correct matches} \\
    \hline
    \multirow{2}{*}{Source} &
    Photo camera &
    72.2 \% \\
    &
    Cellular phone &
    59.3 \% \\
    \hline
    \multirow{2}{*}{Cloud presence} &
    None &
    71.8 \% \\
    &
    Minimal &
    67.9 \% \\
    &
    Massive &
    45.5 \% \\    
    &
    Overcast &
    66.7 \% \\    
    \hline
    \multirow{2}{*}{Skyline composition} &
    Mountains and terrain only  &
    65.1 \% \\
    &
    Foreign objects &
    58.3 \% \\
    \hline    
  \end{tabular}
  \caption{Results of the algorithm with respect to the data set categories of interest.}
  \label{tab:datasetCategoriesResults}
\end{table}

From the results of data set categories it follows that the photographs shot with a photo camera are more frequently aligned than those shot with a cellular phone: an explanation for this trend can be the fact that the photo camera photographers usually use the optical zoom that is correctly annotated in the focal length of the photograph, cellular phones instead usually allow only digital zoom, which leaves unchanged the annotated focal length and so leads to an incorrect Field of View estimation and therefore an incorrect alignment.

The trend of the result as cloud conditions vary follows a logical sense (more clouds leads to a higher number of noise edges and to worse performance) and reveals that the presence of clouds is a significant reason for failure of the algorithm, a problem that will be treated in the next chapter. A good score of the overcast case can be explained by the fact that when the sky is completely covered by clouds there are no edges between the real sky and clouds, so there are no noise edges (it can be thought of as the sky painted by a darker cloud color).

The presence of foreign objects such as trees, buildings, and persons in the skyline obviously penalizes the alignment, but as can be seen from the results it is not a critical issue.

In the next chapter several techniques that should improve the results of mountain identification are presented and discussed.

\subsection{Operating parameter evaluation}
\begin{figure}[h!]
	\centering
	\includegraphics[width=1.00\columnwidth]{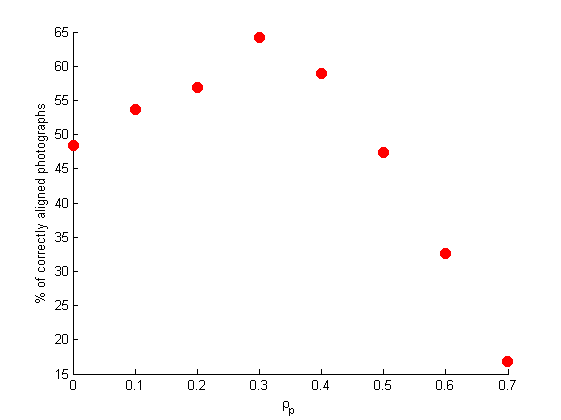}
	\caption{Overall algorithm results varying the $\rho_p$ operating parameter.}
	\label{fig:opRhoP}
\end{figure}

We now present and comment on the effect of the individual parameters on the overall results:

\textbf{\emph{Photograph edge strength threshold ($\rho_p$)}}: The choice of the strength threshold for the photograph edges presents two trends: higher threshold means fewer noise edges, but lower threshold means to include also the weaker edges belonging not to the skyline but to secondary terrain edges. The dependency of the result on $\rho_p$ is displayed in Figure \ref{fig:opRhoP}, the optimal balance between the two trends is reached with $\rho_p = 0.3$.

\begin{figure}[h!]
	\centering
	\includegraphics[width=1.00\columnwidth]{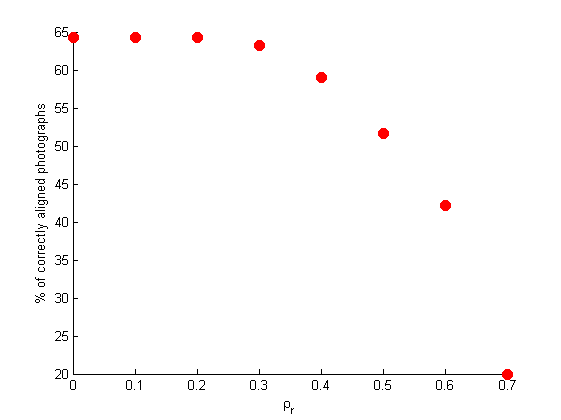}
	\caption{Overall algorithm results varying the $\rho_r$ operating parameter.}
	\label{fig:opRhoR}
\end{figure}

\textbf{\emph{Panorama edge strength threshold ($\rho_r$)}}: The choice of the strength threshold for the panorama edges is very similar to that of the photograph, with a difference regarding small thresholds: since the panorama is rendered by a program it does not present weak edges, which means that the introduction of a small threshold has almost no effect on the matching efficiency. The dependency of the result on $\rho_r$ is displayed in Figure \ref{fig:opRhoR}: in fact all tested values smaller or equal to $0.3$ bring the same (and optimal) result. In the implementation the parameter is set to $\rho_r = 0.2$.

\begin{figure}[h!]
	\centering
	\includegraphics[width=1.00\columnwidth]{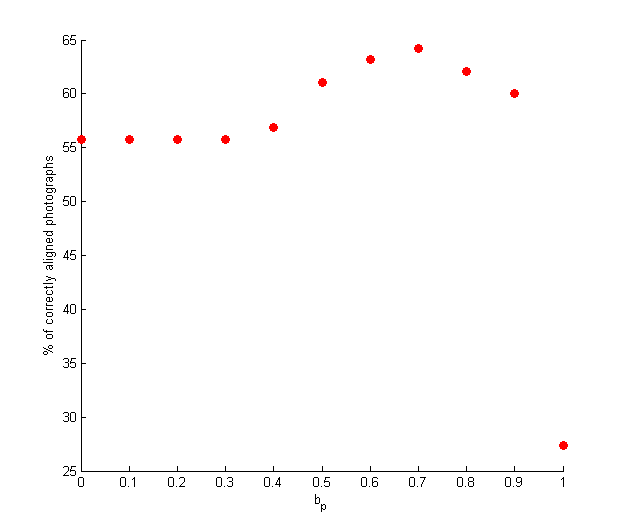}
	\caption{Overall algorithm results varying the $b_p$ operating parameter.}
	\label{fig:opBP}
\end{figure}

\textbf{\emph{Photograph edge filtering base ($b_p$)}}: The choice of the filtering base for the photograph edges defines how strongly the lower edge points will be reduced in strength, $b_p = 1.0$ means that no filtering will be performed, and $b_p = 0.0$ that only the first segment of each row of the photograph will be left. The dependency of the result on $b_p$ is displayed in Figure \ref{fig:opBP}: it is evident that filtering is absolutely needed since the result corresponding to $b_p = 1.0$ is significantly smaller than all the other values; the result stabilizes for $b_p \leq 0.3$ since the base is so small that the filtering factor reaches almost immediately zero. The best orientation is found for $b_p = 0.7$.

\begin{figure}[h!]
	\centering
	\includegraphics[width=1.00\columnwidth]{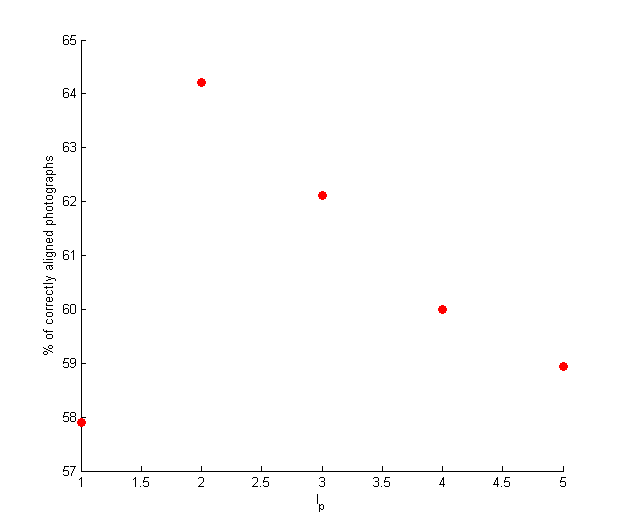}
	\caption{Overall algorithm results varying the $l_p$ operating parameter.}
	\label{fig:opLP}
\end{figure}

\textbf{\emph{Photograph edge filtering maximum segment length ($l_p$)}}: The choice of the maximum segment length for the photograph edge filtering is introduced to allow mountain edges which are thicker than one pixel not to penalize the other edges placed below it, but at the same time to not allow the noise edges that generate columns of points to be left unfiltered. The dependency of the result on $l_p$ is displayed in Figure \ref{fig:opLP}: the optimal balance between the two trends is reached with $l_p = 2$.

\textbf{\emph{Panorama edge filtering base ($b_r$)}}: The choice of the filtering base for the panorama edges defines how strongly the lower edge points will be reduced in strength, $b_r = 1.0$ means that no filtering will be performed, and $b_r = 0.0$ that only the first segment of each row of the photograph will be left. The dependency of the result on $b_p$ is displayed in Figure \ref{fig:opBP}: the overall trend is similar to the filtering base of photograph edges ($b_p$) but with an important difference for high values: the result does not get worse with $b_r$ getting close to $1$. This can be explained by the fact that the high values ($\geq 0.8$) of $b_p$ do not perform a significant filtering, so the noise edge points are not filtered and penalize the matching: the panorama edges instead do not have any noise, so do not have the same trend. The choice to set $b_r = 1.0$ is very clear, or in other words to not perform any filtering of panorama edge points.

\begin{figure}[h!]
	\centering
	\includegraphics[width=1.00\columnwidth]{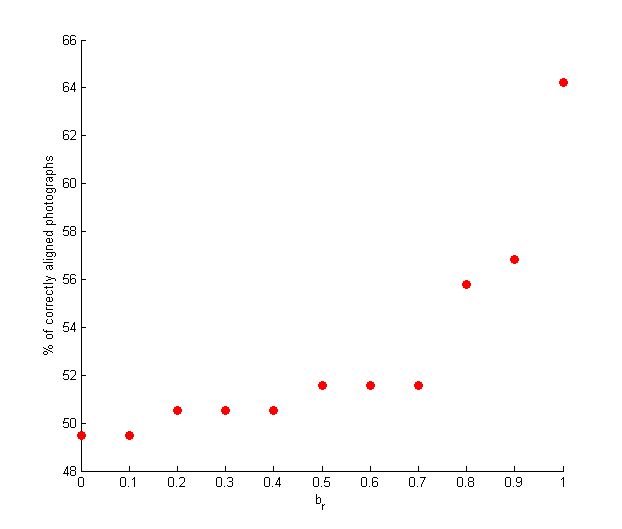}
	\caption{Overall algorithm results varying the $b_r$ operating parameter.}
	\label{fig:opBR}
\end{figure}

\textbf{\emph{Panorama edge filtering maximum segment length ($l_r$)}}: since $b_r$ is set to $1$ (the filtering of the panorama edges is not performed) the maximum segment length for the panorama edges filtering has no impact on the result, so its choice is not significant.

\begin{figure}[h!]
	\centering
	\includegraphics[width=1.00\columnwidth]{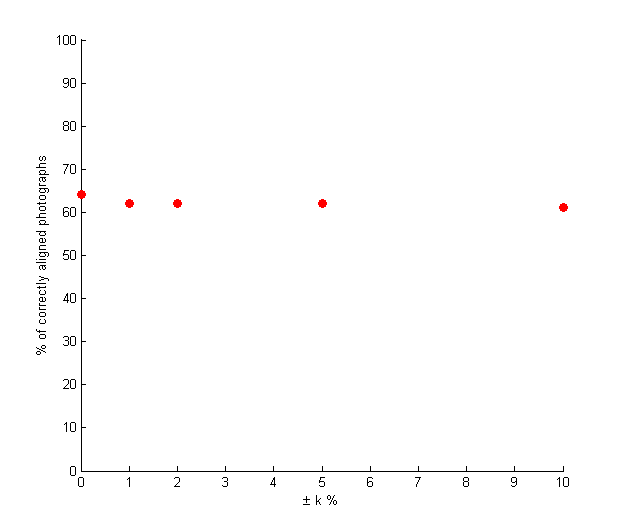}
	\caption{Overall algorithm results varying the $\pm k \%$ operating parameter.}
	\label{fig:opK}
\end{figure}

\textbf{\emph{Photograph scaling interval ($\pm k \%$)}}: The choice of the scaling interval with respect to the originally estimated scaling process is described in the previous chapters. The results of its estimation were probably the most unexpected of all operating parameters. The dependency of the result on $b_p$ is displayed in Figure \ref{fig:opK}: the best performance is reached with $k = 0$ (no scaling interval, only the estimated scale factor) and always worsens with increasing $k$. This trend can be interpreted as the proof of the fact that the Field of View and scaling factor estimation are so precise that increasing the scaling interval brings more penalties by setting an imprecise scale factor and increasing the probability of a noise edge matching the mountain panorama edge than advantages by trying different zoom factors and getting the best mountain edge overlap. The best orientation is therefore found with $k = 0$ (no scaling interval).

\chapter{Conclusions and Future Work}
\label{Conclusions and Future Work}
\thispagestyle{empty}

\vspace{0.5cm}

\noindent
In this work an algorithm for the estimation of the orientation of a geo-tagged photograph containing mountains and for the identification of the visible mountain peaks is presented in detail. The algorithm has been implemented and evaluated with the result of estimating correctly the direction of the view of \textbf{64.2\%} of the various photographs.

This result can be considered excellent if used in passive crowdsourcing applications: given the enormous availability of photographs on the Web, the desired amount of correctly matched photographs can be reached just by increasing the number of processed photographs.

It is a very promising even if not excellent result speaking about applications connected with user experience: approximately one photograph out of three may not be matched, decreasing significantly the usability of the application.

\section{Future enhancements}
We expect to implement and test several techniques in the near future, aimed at enhancing the matching process and improving the results.

\begin{figure}[h!]
	\centering
	\includegraphics[width=1.0\columnwidth]{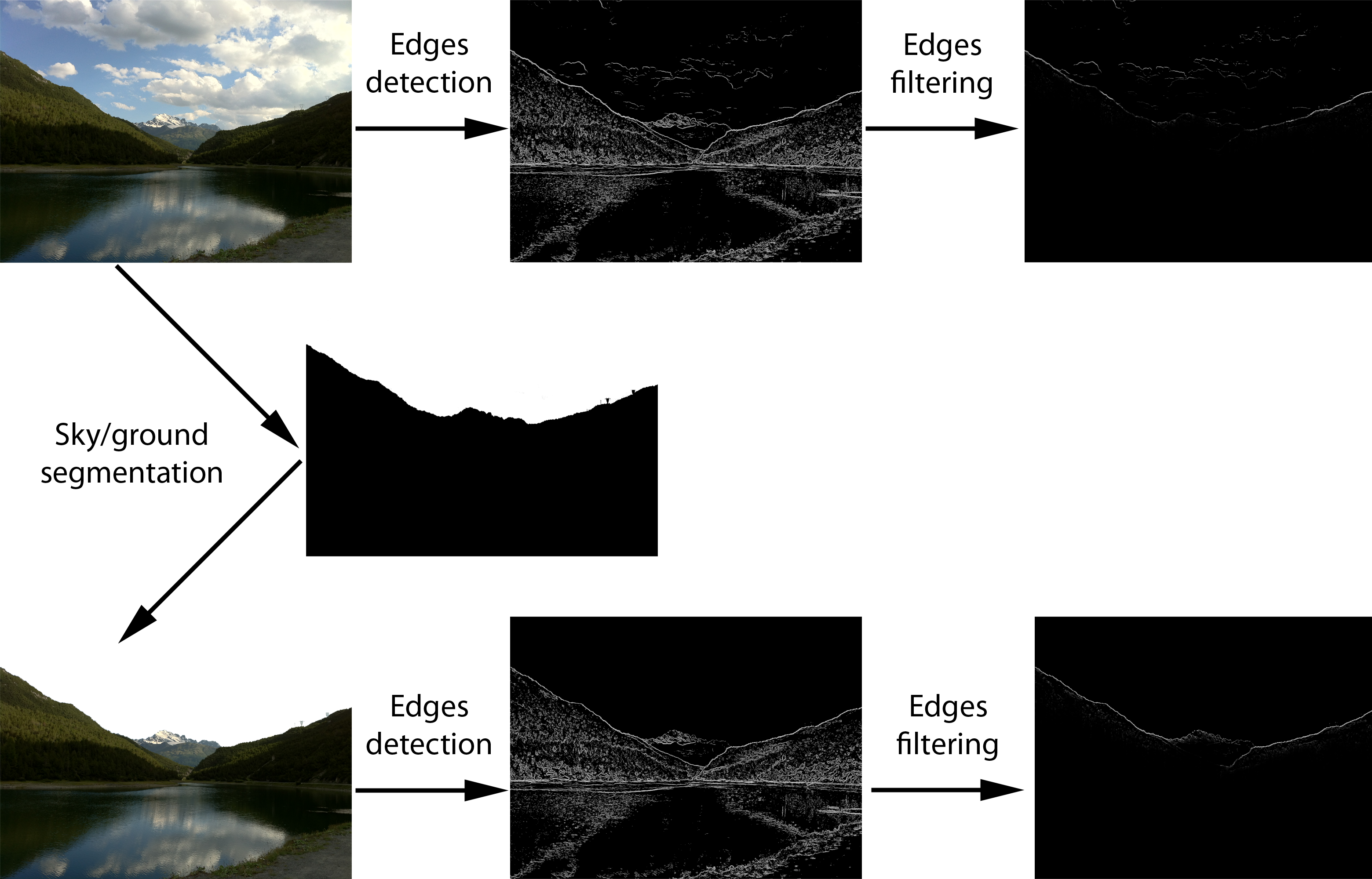}
	\caption{An example of the use of sky/ground segmentation in the edge extraction technique: noise cloud edges are removed and the mountain boundary is perfectly extracted.}
	\label{fig:skyTerrain}
\end{figure}

One of the main reasons that the matching fails is the massive presence of clouds on the photograph, and to prevent the edge filtering step from emphasizing the noise cloud edges a sky/ground segmentation step can be inserted before the edge extraction phase: a sky/ground segmentation algorithm (such as, for example, that proposed by Todorovic and Nechyba \cite{Todorovic03sky/groundmodeling}) that given a photograph splits it into the two regions of sky and terrain, then the sky part is erased from the photograph and the matching algorithm goes on without the cloud edge noise. In Figure \ref{fig:skyTerrain} an example of this approach and the improvement it brings to the filtered edges map is shown.

\begin{figure}[h!]
	\centering
	\includegraphics[width=1.0\columnwidth]{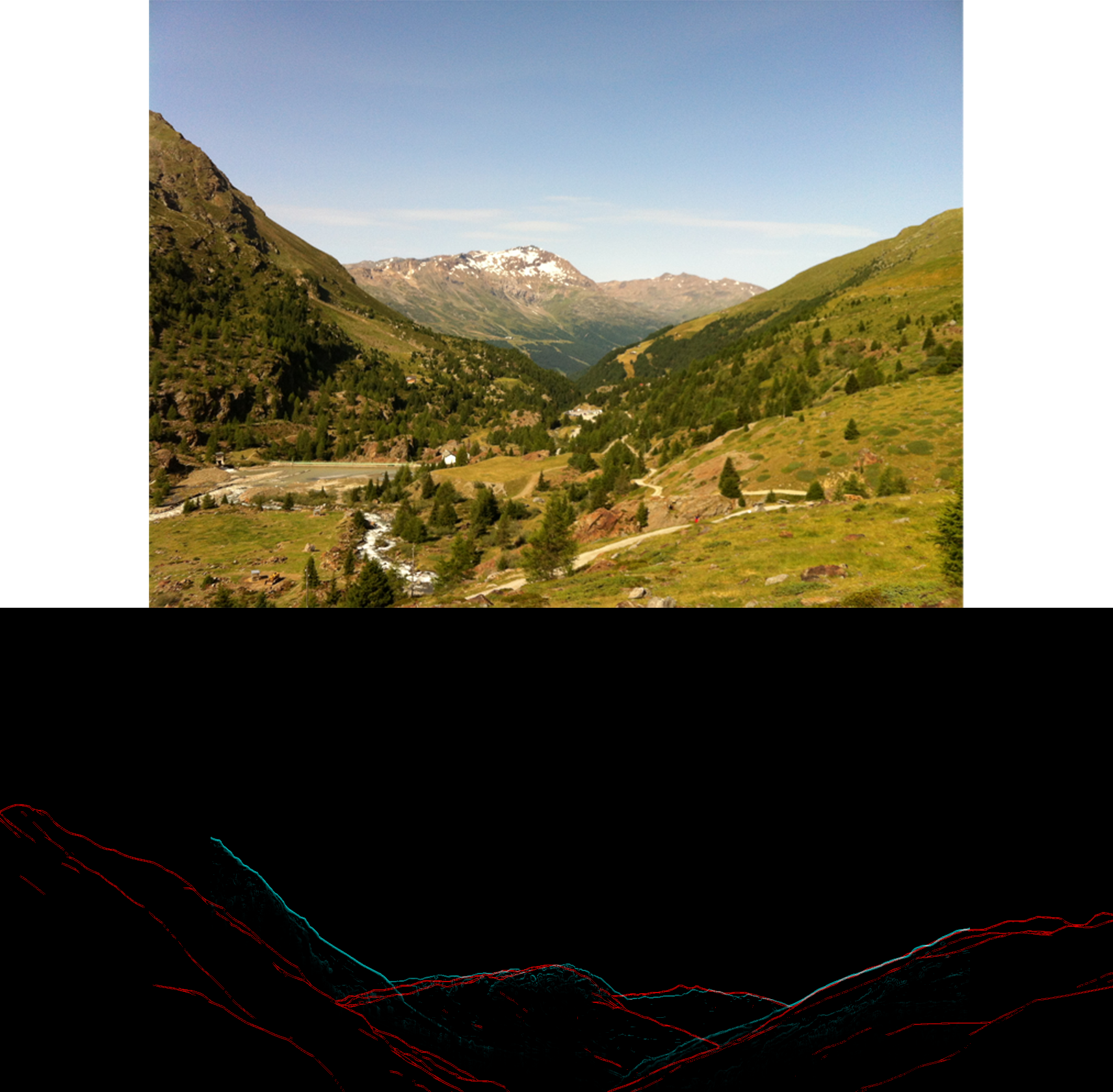}
	\caption{An example of a photograph (top) of background mountains seen in the aperture between the foreground mountains and the corresponding edge matching (bottom, red - panorama, blue - photograph).}
	\label{fig:viewAperture}
\end{figure}

Another frequent reason for incorrect matching is the imperfections between the boundaries on the photograph and on the render of mountains very close to the observer. Due to the small and unavoidable errors in the registered position, altitude estimation, and elevation model, the rendered panorama will always be imperfect with respect to reality, and this imperfection, for obvious reasons, get smaller as the distance from the observer to the object increases. The situation of having mountain boundaries both in the foreground and the background is very frequent: walking in a mountainous area the observer is usually surrounded by mountains placed close to him that are obscuring the distant mountains; the dimensions and the majesty of the mountains however bring usually the photographer to take pictures of distantly placed mountains with altitude higher that his point of view. For this reason as soon as the mountains in the foreground get placed in a way to create an "aperture", a photograph of the mountains visible in this sort of window will be probably shot. An example of this type of photograph can be that used for Figure \ref{fig:skyTerrain} or that shown in Figure \ref{fig:viewAperture}: it represents exactly the described situation, and it can be easily seen that the photograph and panorama edges of the mountain in the background are perfectly matchable, while the edges belonging to the closer mountains placed in the foreground are significantly different. In this case the algorithm manages to correctly alignment of the photograph anyway, but it is frequent in these cases of funnel-shaped mountain apertures that it fail, specially if the aperture is very small with respect to the total photograph dimension. Future studies are planned to include the development of a technique for the recognition of this type of situation and emphasizing the edges of the background mountains with respect to the foreground mountains.

Even when the estimated orientation is not correct, in most cases the correct alignment presents a significant peak in the VCC score distribution, so the best edge overlap estimation could be improved with a robust matching technique operating on the top-N positions extracted from the VCC matching score distribution. The likely method of implementing it is the neighborhood metrics of the edges as proposed by Baboud et al. \cite{Baboud2011Alignment}; even if a simplified implementation of the robust matching technique proposed was implemented in this work and rejected as self-defeating, we believe that it can be refined to reach a higher rate of correctly matched photographs.

\section{Crowdsourcing involvement}
An important future direction of the research of this work is the integration with (traditional) crowdsourcing. Though being a feasible image processing task, photograph-to-panorama alignment is definitely better performed by humans that by algorithms. There is no need to move the photograph through the panorama or know the correct scale factor: a person can usually easily align the photograph to the panorama just by looking at it, comparing the fragments of the photograph he finds most particular, which can be elevated peaks, details that catch the eye and in general sense any terrain silhouette fragments that the person considers unusual. This allows one to easily identify the correct alignment by eye even in presence of significant errors between the photograph and the generated panorama.

This consideration leads us to the integration of the current work with crowdsourcing with several possible scenarios:
\begin{itemize}
\item
\emph{Contribution in learning and testing phase}:
manual photograph-to-panorama matching estimation can be very useful as a source of ground truth data both for the learning phase in case of using a machine intelligence algorithm in the edge matching and for the implementation testing phase, validating the results proposed by the algorithm. These approaches can be applied both to expert and non-expert crowdsourcing options: the non-expert tasks can consist of an activity feasible by anyone, such as searching for the correct alignment between the photograph and the corresponding panorama, the expert tasks instead aim at the activities that only mountaineers can perform, such as tagging the mountain peaks on a photograph based on his own knowledge and experience. Both methods allow the validation of the algorithm estimation result, but at two different (even if very related) levels: the first at photograph direction estimation, and the second at final mountain peak identification and tagging.
\item
\emph{Contribution as post processing validation}:
human validation can be used not only in the development phase but also in the final implementation pipeline as the post processing validation. If the application is time-critical such as real-time augmented reality or website photographs mountain peak tagging, this approach cannot be applied, but in crawler applications such as environmental model creation, crowdsourcing can provide a significant improvement in the data quality by proposing manual photograph-to-panorama matching for the photographs that the algorithm has marked as photographs with low confidence score. Crowdsourcing therefore will be used as a complementary technique for photographs that cannot be aligned automatically.
\item
\emph{Contribution to data set expansion}:
the number of the available photographs in the data set is fundamental both for testing purposes and for the data to be processed by an environmental modeling system. Photographs can be collected from public sources such as social networks or photo sharing websites, filtering the geographical area and image content of the interest, or can be collected directly from people, uploading or signaling the photographs containing mountains in a certain requested area, or even photographs containing certain requested mountain peaks. This approach is important in the case of environmental modeling, when precisely collected photographs in the same area of ground truth data availability is fundamental.
\end{itemize}

\section{Possible areas of application}
There are several application areas the proposed algorithm can be used for, starting from the technique for snow level monitoring and prediction (the technique this work has been proposed and started for) and extending to possible applications in mobile and web fields that can be created thanks to this algorithm.

\subsection{Environmental Modeling}
One of the main future directions of the research will be the modeling of environmental processes by the analysis of mountain appearances extracted by the algorithm proposed in this work. The most important and most obvious measurements available from a mountain's appearance are the snow level and the snow water equivalent (SWE). The idea is to collect a series of photographs through time for each analyzed mountain, and based on the snow data ground truth, estimate the current measurements. The first step will therefore be an attempt to detect the correlation between the visual content of the mountain portion of a photograph with the physical measurements of the snow and SWE of that mountain. An interesting planned approach is to exploit also webcams in the region of interest since they present several advantages:
\begin{itemize}
\item
weather and tourist webcams are very popular and frequent in mountain regions
\item
the time density of the measures can be as high as we want (it is only a matter of how frequently the image is acquired from the webcam, and in any case there is no reason to suppose the need of more than a couple of captures per day)
\item
the mountains captured on a webcam are always in the same position on the image, so even in cases of difficult edge matching it can be done or verified manually once to exploit all the instances of the photographs in future.
\end{itemize}

Furthermore, supposing the mountains to be greatly distant from the observer (a reasonable and weak assumption in case of mountains photographs), by knowing the estimation of the observer's altitude and the altitude of the identified peak, we can estimate the altitude of each pixel of the mountain on the photograph by a simple proportion of the differences of altitudes and the pixel height of the mountain. This allows a comparison between the visual features of partial photographs corresponding to the same altitude even for photographs with significantly different shot position and photo camera properties.

\subsection{Augmented Reality}
Augmented reality applications on mobile devices (applications that augment and supply the real-time camera view of the mobile device with computer-generated input) is a recent niche topic, and the promising use of the described algorithm regards augmented reality: an application can tag in real-time the mountains viewed by the user, highlight the peaks and terrain silhouettes, and augment the mountains with useful information such as altitude contours drawn on the image.

Such a kind of application would eliminate the problem of wrong geo-tag estimation (keeping the GPS of the mobile device on, the position is usually estimated with a tollerance of few meters) and so will reduce significantly the problem of wrong altitude estimation and elevation model imperfections with respect to reality. The reduced computation capacity may be compensated by the built-in compass, which gives a rough indication of the observer's direction of view, so the matching procedure can be done only on a reduced fragment of the rendered panorama. The bandwidth use will be small since the mountains are usually distant from the observer, so the rendered view will change very slowly while the observer is moving, so it will need to be updated rarely.

\subsection{Photo Sharing Platforms}
Tagging the mountain peaks on the geo-tagged photographs can lead also to a significant improvement to a cataloging and searching system of a social network or a photo sharing website, and to a better user experience by exploring the peak names and other information by directly viewing the photograph.
Automatic tagging of mountain peaks (tagging intended as the catalog assignment of peak names to the photograph and not the visual annotation on the photograph itself) can allow navigation through mountain photographs in these scenarios:
\begin{itemize}
\item
User searches for a mountain by specifying its name in the query: retrieves the photographs of that mountain, even if the author of the photograph did not specify the name in the title, description or other metadata.
\item
User views a photograph: other photographs of the same mountain (with the same or different facade) are suggested, even if the authors of both photographs did not specify the name in the title, description or other metadata.
\end{itemize}

\cleardoublepage
\addcontentsline{toc}{chapter}{Bibliography}
\bibliographystyle{plain}
\bibliography{bib}

\appendix

\pagestyle{fancy} 
\fancyfoot{}                                               
\renewcommand{\chaptermark}[1]{\markboth{\appendixname\ \thechapter.\ #1}{}} 
\renewcommand{\sectionmark}[1]{\markright{\thesection.\ #1}}         
\fancyhead[LE,RO]{\bfseries\thepage}    
                                        
\fancyhead[RE]{\bfseries\leftmark}    
\fancyhead[LO]{\bfseries\rightmark}     
\renewcommand{\headrulewidth}{0.3pt}

\end{document}